\documentclass{article}

\usepackage[preprint]{neurips_2026}

\usepackage[utf8]{inputenc} 
\usepackage[T1]{fontenc}    
\usepackage{hyperref}       
\usepackage{url}            
\usepackage{booktabs}       
\usepackage{amsfonts}       
\usepackage{nicefrac}       
\usepackage{microtype}      
\usepackage{xcolor}         
\usepackage[normalem]{ulem} 

\usepackage{amsmath}
\usepackage{amssymb}
\usepackage{mathtools}
\usepackage{amsthm}

\usepackage{microtype}
\usepackage{graphicx}
\usepackage{subcaption}
\usepackage{wrapfig}

\usepackage{hyperref}

\usepackage{fontawesome} 
\usepackage{multirow} 
\usepackage{booktabs}
\usepackage{array}
\usepackage{makecell} 
\usepackage{float}  

\title{THEIA: A Multimodal Dataset and Benchmark for Vision-Language Analysis of Layout}

\author{%
	Giuseppe Chiari \\
	DEIB\\
	Politecnico di Milano\\
	Milan, Italy, 20133\\
	\texttt{giuseppe.chiari@polimi.it} \\
	\And
	Michele Piccoli \\
	DEIB \\
	Politecnico di Milano \\
	Milan, Italy, 20133 \\
	\texttt{michele.piccoli@polimi.it} \\
	\And
	Federico Viola \\
	DEIB \\
	Politecnico di Milano \\
	Milan, Italy, 20133 \\
	\texttt{federico.viola@polimi.it} \\
	\And
	Davide Zoni \\
	DEIB \\
	Politecnico di Milano \\
	Milan, Italy, 20133 \\
	\texttt{davide.zoni@polimi.it} \\
}

\begin{document}

\maketitle

\begin{abstract}
	The integration of artificial intelligence into computer-aided design frameworks has sparked a shift in the design of analog integrated circuits~(ICs), transitioning the field from using manual and algorithmic-based solutions to adopting automated and intelligent paradigms. In this scenario, the GDSII file represents the industry-standard database containing the ultimate and most accurate source of information of the analog circuit, encapsulating the complex physical geometries and parasitic realities that define tape out performance. This paper proposes THEIA, a novel dataset containing thousands of layout images paired with question-answer conversations, along with a benchmark that employs a fine-tuned vision-language model~(VLM) to analyze GDSII files of analog circuits, enabling designers to interact with and query physical layouts as intuitive, meaningful entities. Experimental results using thousands of analog designs across five realistic tasks demonstrate that the proposed fine-tuned VLM outperforms state-of-the-art general-purpose VLMs by a significant margin~(up to $73\%$), highlighting a fundamental gap between general-purpose multimodal reasoning and domain-specific layout understanding.
\end{abstract}

\section{Introduction}
\label{sec:introduction}
\begin{figure}[t]
	\centering
	\begin{minipage}[b]{0.48\textwidth}
		\centering
		\includegraphics[width=\linewidth]{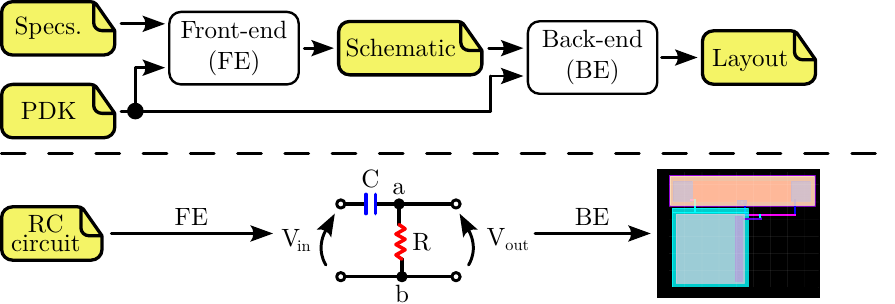}
		\caption{Overview of the analog design flow~(top). Design specifications and target process design kit~(\texttt{PDK}) are first translated into a \texttt{Schematic} that captures the required functionality by the \texttt{Front-end} step. The schematic is then realized as a physical \texttt{Layout} through the \texttt{Back-end} stage. A simplified example~(bottom) is provided to highlight the two design stages.}
	\label{fig:eda_flow}
	\end{minipage}
	\hfill
	\begin{minipage}[b]{0.48\textwidth}
		\centering
		\includegraphics[width=\linewidth]{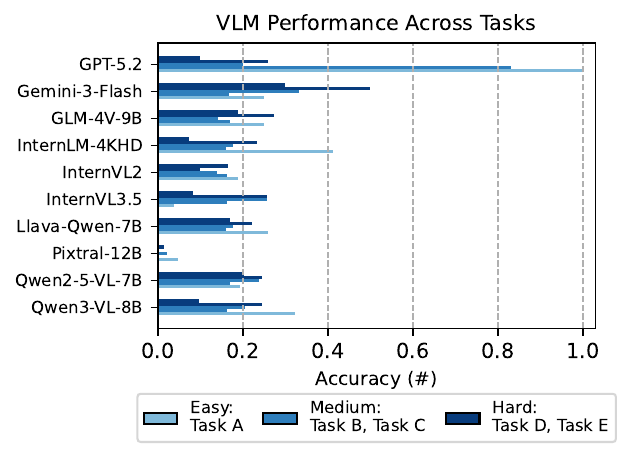}
		\caption{Performance overview of off-the-shelf VLMs on analog layout analysis tasks, highlighting poor accuracy across models.}
		\label{fig:motivational_example}
	\end{minipage}
\end{figure}

Traditionally, the analog design flow is a manual, heuristic-driven and extremely time-consuming process, due to the fact that it heavily relies on the intuition of senior analog engineers to navigate complex trade-offs between gain, bandwidth, and power. As depicted in Figure~\ref{fig:eda_flow}, the analog design flow is generally seen as organized into two distinct macro-phases: \emph{(i)}~the front-end, which uses the schematic representation of the circuit to perform topology selection, transistor sizing and pre-layout simulations,  and \emph{(ii)}~the back-end, which uses the GDSII representation to perform the physical implementation of the design through manual layout, parasitic extraction, and final verification.
While the integration of artificial intelligence~(AI) into analog computer-aided design~(CAD) frameworks has recently sparked a seminal shift in this domain, transitioning the field from manual, iterative tuning toward automated, intelligent optimizations the advance is mainly confined to front-end tasks. This is due to the complexity of accessing and taking advantage of the extremely detailed source of information within the GDSII file of the circuit.
Notably, the Graphic Data System II~(GDSII) is the industry-standard database for the exchange of integrated circuit layout, which contains all the information to analyze, verify, optimize, and manufacture the circuit.
While schematic-level simulations provide a theoretical baseline for the behavior of the circuit, the GDSII file captures the actual behavior of the circuit in the form of parasitics, e.g., electromigration risks, substrate coupling, and unintended capacitive effects, that ultimately determine whether the manufactured chip implementing the circuit will meet  specification performance and requirements. 
However, extracting high-level semantic insights from GDSII files, such as identifying the total device count or detecting specific sub-circuit topologies, requires exhaustive manual inspection and the execution of rigid, computationally expensive, and rule-based scripts~\cite{XZL+2019, KMS+2019}. Moreover, with the increasing design complexity the ability of human designers to manually parse, verify, and check these geometric representations is becoming increasingly unsustainable. 
To this end, it is paramount to start perceiving the GDSII as a multi-layered, high-resolution image, 
to introduce the possibility of applying automated computer vision techniques bridging the gap between raw geometric polygons representation and circuit-level intents. 
Deep learning-based detectors have been successfully applied across diverse domains, including construction~\cite{LFH+2021}, cybersecurity~\cite{CGL+2024,GCZ2024,GCZ2025}, and autonomous driving~\cite{CLL+2021}.
Building on the success of AI assistants in software engineering and digital RTL design~\cite{BGK+2023, CWR+2023} and the emergence of vision language models~(VLMs), we foresee a unique opportunity to introduce a novel, semantic-based approach to the use of GDSII in the analog design flow. VLMs, which combine the perceptual power of computer vision with the reasoning capabilities of large language models~(LLMs), have demonstrated remarkable proficiency in interpreting natural images and technical diagrams. However, their application to the semiconductor domain is challenging and, to the best of our knowledge, never explored before for two reasons. First, off-the-shelf VLMs are trained using datasets, e.g., photographs of landscapes, objects, and people, which share almost nothing with the abstract, multi-layered, and highly regulated geometries of a GDSII layout thus exhibiting limited performance on analog layout analysis tasks. Indeed, Figure~\ref{fig:motivational_example} reports limited performance for different vanilla VLMs when requested to perform five identification tasks targeting a set of GDSII files: single-device identification~(Task A), recognition of simple circuit topologies~(Task B), identification of complex topologies~(Task E) and counting of devices in both simple~(Task C) and complex~(Task D) layouts. Notably, the commercial models equipped with reasoning support, such as GPT 5.2, perform well on simple tasks, while the quality of the results strongly decreases on medium and complex tasks, e.g., Tasks C, D, and E.
Second, the lack of specific training datasets due to the use of manual effort to create and validate each circuit layout severely limits the use of automated AI-powered solutions. 

This work introduces THEIA, a multimodal dataset of analog layout images paired with question-answer conversations, designed to advance automated understanding of GDSII layouts and enable interactive, AI-assisted analysis. By treating physical layouts as visual entities rather than collections of polygons, THEIA enables intuitive querying of layout structures through natural language, moving from passive verification toward interactive analysis. In addition, this work introduces a benchmark methodology using domain-specific fine-tuning of a pretrained vision-language model, showing that general-purpose VLMs struggle to interpret analog layouts without adaptation.
This setup is evaluated on five realistic tasks spanning multiple circuit topologies and complexity levels. Results show that domain adaptation leads to substantial performance gains, with fine-tuned models outperforming strong zero-shot baselines by up to 73\% accuracy. In particular, this work delivers three contributions to the state of the art:
\begin{itemize}
	\item \textbf{Open dataset and end-to-end workflow.} THEIA, a curated multimodal dataset~(layout images paired with question-answer conversations) together with an open-source training and evaluation pipeline, enabling reproducible VLM fine-tuning and benchmarking.\footnote{Repository: \href{https://huggingface.co/datasets/hardware-fab/theia}{hardware-fab/theia}}
	The released tasks span increasing difficulty, from single-device identification to base-topology recognition and more complex reasoning over full circuit layouts.
    \item \textbf{Task-specific VLM adaptation.} A fine-tuned VLM specialized for analog-layout understanding, paving the way for a new generation of AI-augmented EDA tools and the transition of the analog design field from manual, iterative tuning toward automated, intelligent optimization.
    \item \textbf{Experimental validation on real layouts.} We perform an extensive evaluation on a large collection of representative analog circuits using DRC-free and LVS-compliant GDSII layouts, and we compare against state-of-the-art VLM baselines. 
\end{itemize}
The remainder of the paper is organized into five sections. 
Section~\ref{sec:background} reviews related work on AI-driven analog design automation. 
Section~\ref{sec:methodology} describes the dataset creation process and proposed VLM fine-tuning and deployment methodology. 
Section~\ref{sec:results} reports the experimental evaluation, and
Section~\ref{sec:conclusions} concludes the paper with final remarks.
The Appendix~\ref{sec:a_appendix} provides additional implementation, dataset, and evaluation details, including fine-tuning and deployment configurations, dataset statistics and layout examples, task records examples, and answer-option construction.

\section{Related Work}
\label{sec:background}
\noindent\textbf{Front-end.} Several works focus on schematic analysis. Netlistify~\cite{HCH+2025} employs a hybrid approach combining convolutional neural networks~(CNNs) and transformers to identify devices, orientations, and connections within analog circuit schematics. In the realm of topology generation, LaMAGIC~\cite{CYS+2024} fine-tunes LLMs specifically for power converters by developing structured input-output representations. Similarly, AnalogGenie~\cite{GCY2025} utilizes a GPT-based model trained on a large dataset to predict sequential pin connections for various analog topologies. For code-centric approaches, AnalogCoder~\cite{YSG+2024} proposes a training-free LLM method for Python-based circuit generation, while FLAG~\cite{MYT+2024} enhances LLMs with formulaic knowledge to assist in baseband circuit design, mitigating the need for extensive programming expertise. 
Beyond circuit topology, LLMs are increasingly deployed as agents for parameter sizing and verification. LADAC~\cite{Liu_2024} introduces an agent incorporating a knowledge library to assist in transistor sizing and simulation. To enhance reasoning capabilities, Artisan~\cite{CHL+2024} employs tree-of-thought~(ToT) and chain-of-thought prompting for operational amplifier design. AnalogXpert~\cite{HSY+2025} further streamlines synthesis by incorporating a subcircuit library and iterative proofreading. Optimization efficiency is addressed by ADO-LLM~\cite{YWX+2024}, which combines Bayesian Optimization with in-context learning, and LEDRO~\cite{DHA+2025}, which uses LLMs to refine search regions for existing optimization methods.
\begin{figure}[t]
	\centering
	\includegraphics[width=0.9\linewidth]{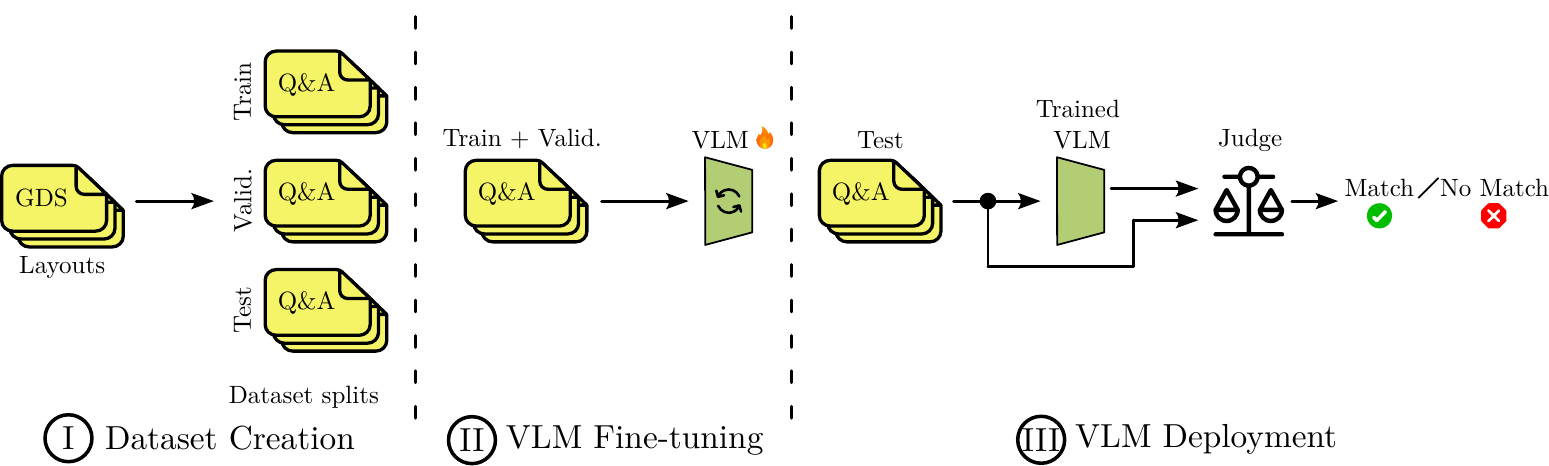}
	\caption{Methodology overview. It comprises three main phases: \emph{(i)}~\textit{Dataset Creation}, \emph{(ii)}~\textit{VLM Fine-tuning}, and \emph{(iii)}~\textit{VLM Deployment}. \textit{Dataset Creation} generates sets of task-specific questions and answers~(\texttt{Q\&As}) from a collection of GDSII layouts~(\texttt{GDS}). \textit{VLM Fine-tuning} uses training and validation splits to fine-tune a VLM to the target domain. \textit{VLM Deployment} evaluates the fine-tuned model on the held-out test data, employing an LLM-as-a-judge to assess prediction correctness.}
	\label{fig:flow}
\end{figure}

\noindent\textbf{Back-end.} Recent research has begun to address the complexities of physical layout and verification using generative models without directly addressing the complexity of GDSII analysis. ACDC~\cite{YJA+2025} frames transistor placement as a sequence prediction problem, fine-tuning transformers to predict coordinates directly from a netlist. SOLOMON~\cite{BX2025} demonstrates the adaptation of general-purpose LLMs for semiconductor layout by utilizing ToT reasoning to generate Python scripts for complex 3D structures.
Regarding verification, LLM-HD~\cite{CWW+2024} introduces a layout language model that processes GDSII binaries to detect lithography hotspots without image conversion. DRC-Coder~\cite{CHL+2025} proposes a multi-agent framework to interpret textual design rules and generate standard verification rule format (SVRF) code. However, these works focus on placement prediction and verification rather than layout analysis and identification.
Other works target the layout of analog circuits. BAG~\cite{CPL+2013,CHB+2018} introduced parameterized generators for analog and mixed-signal circuits, enabling automated schematic and layout generation but relying heavily on predefined templates and domain expertise. ALIGN~\cite{KMS+2019} and MAGICAL~\cite{XZL+2019} employ an algorithmic approach. Recent works proposed AI-driven layout methodologies using DNN~\cite{BBL+2021}, Bayesian~\cite{BZC+2023,GZY+2024}, reinforcement learning~\cite{BBV+2024,BBV+2025a,DBB+2025,CPZ2026}, and graph neural network~\cite{BBV+2025a} models.
Notably, the use of AI-driven methodologies to improve the analog design flow is blooming, while, to the best of our knowledge, this is the first proposal that directly targets the GDSII representation of analog circuits, paving the way for the emergence of analog AI-assistants.

\section{THEIA Dataset Creation and Benchmark Methodology}
\label{sec:methodology}
This section presents the benchmark methodology for fine-tuning a VLM to analyze GDSII layouts of analog circuits, enabling task-specific question answering and semantic interpretation of layout structures. Figure~\ref{fig:flow} illustrates the overall workflow, which is organized into three main stages: \emph{(i)} \textit{Dataset Creation}, \emph{(ii)} \textit{VLM Fine-tuning}, and \emph{(iii)} \textit{VLM Deployment}. Starting from a collection of GDSII layouts, the proposed methodology first constructs a dataset tailored to the requirements of VLM training, subsequently fine-tunes the model, and finally deploys and evaluates it.
%
\begin{figure*}[t]
	\centering
	
	\begin{subfigure}{\textwidth}
		\includegraphics[width=\linewidth]{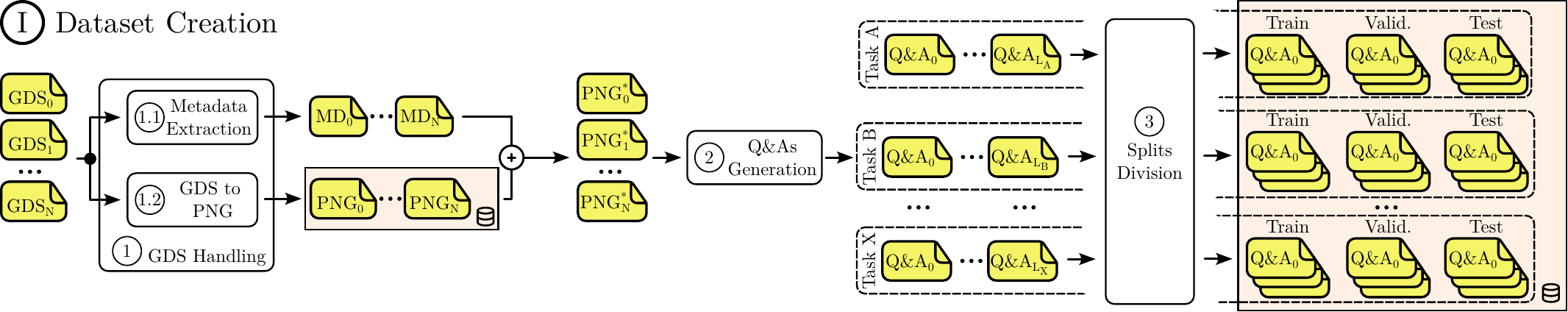}
		\caption{\textit{Dataset Creation} stage. Starting from a collection of analog layouts in GDSII format, the pipeline generates task-specific training, validation, and test splits through three sequential steps: \emph{(i)} \textit{GDS Handling}, \emph{(ii)} \textit{Q\&As Generation}, and \emph{(iii)} \textit{Splits Division}. \textit{GDS Handling} renders each GDSII file into a high-level layout representation. \textit{Q\&As Generation} produces task-dependent questions and corresponding ground-truth answers. Finally, \textit{Splits Division} organizes the resulting \texttt{Q\&As} into task-specific sub-datasets.}
		\label{sfig:dataset_creation}
	\end{subfigure}
	
	\begin{subfigure}{\textwidth}
		\centering
		\includegraphics[scale=0.55]{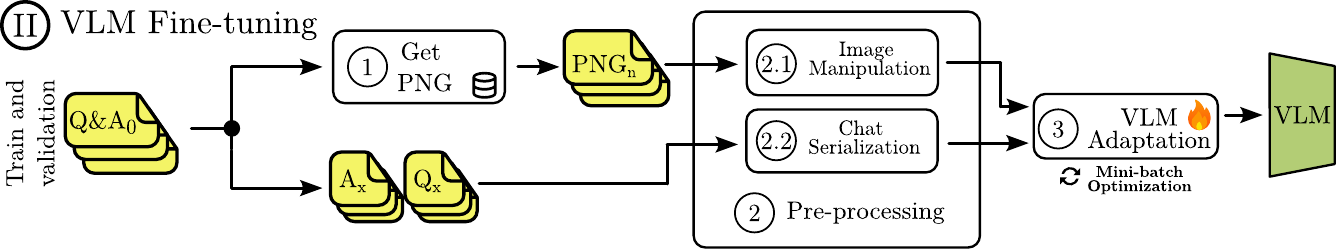}
		\caption{\textit{VLM Fine-tuning} stage. Task-specific training and validation \texttt{Q\&As} are used to adapt a pre-trained VLM through three sequential steps: \emph{(i)}~\textit{Get PNG}, \emph{(ii)}~\textit{Pre-processing}, and \emph{(iii)}~\textit{VLM Adaptation}. \textit{Get PNG} retrieves the layout image $\texttt{PNG}_\texttt{n}$ associated with each \texttt{Q\&A}. \textit{Pre-processing} prepares both visual and textual inputs in the model-specific format, while \textit{VLM Adaptation} fine-tunes the model.}
		\label{sfig:vlm_finetune}
	\end{subfigure}
	
	\begin{subfigure}{\textwidth}
		\centering
		\includegraphics[scale=0.55]{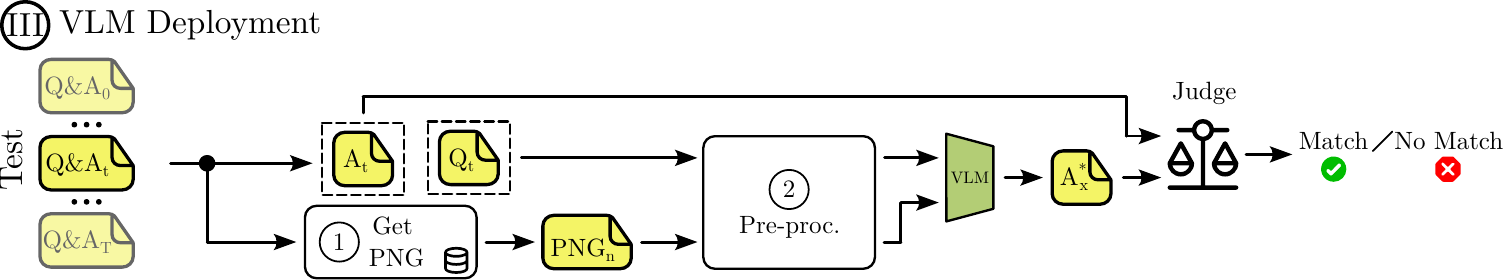}
		\caption{\textit{VLM Deployment} stage. The fine-tuned VLM is evaluated on task-specific test samples processed individually. For each \texttt{Q\&A}, the layout image and pre-processed question are provided as input to the VLM, which generates a predicted answer. An LLM-as-a-judge then performs a semantic comparison between the predicted and ground-truth answers, producing a binary outcome for evaluation.}
		\label{sfig:vlm_deployment}
	\end{subfigure}
	\label{fig:vlm_pipeline}
	\caption{Detailed methodology stages: \emph{(i)}~\textit{Dataset Creation}~(top), \emph{(ii)}~\textit{VLM Fine-tuning}~(middle) and \emph{(iii)}~\textit{VLM Deployment}~(bottom).}
\end{figure*}

\subsection{THEIA Dataset Creation}
\label{ssec:dataset_creation}
The first stage focuses on organizing the data required for subsequent VLM training. Figure~\ref{sfig:dataset_creation} illustrates the pipeline adopted to generate the dataset.
A collection of real GDSII analog layouts is generated from human-designed circuit templates using the OSIRIS open-source layout generator~\cite{CPZ2026}, with device sizing and number of fingers varied. Design rules checking~(DRC) is performed by the generator, and Netgen layout-vs-schematic~(LVS) verification confirms correspondence to the source netlists. All layouts used in THEIA pass both checks.
In particular, the \emph{Dataset Creation} phase consists of three sequential steps: \emph{(i)}~\textit{GDS Handling}, \emph{(ii)}~\textit{Q\&As Generation}, and \emph{(iii)}~\textit{Splits Division}.
The \textit{GDSII Handling} step is responsible for extracting relevant information from each GDSII layout and consists of two sub-steps: \emph{(i)} \textit{Metadata Extraction} and \emph{(ii)} \textit{GDSII to PNG}. The former, \textit{Metadata Extraction}, sub-step extracts descriptive attributes for each layout, including the device type, the set of devices present, and their corresponding counts from the GDSII layout file. For each input layout~($\texttt{GDS}_\texttt{n}$), this process produces a metadata file~($\texttt{MD}_\texttt{n}$) summarizing its key characteristics. In parallel, the latter sub-step,  \textit{GDS to PNG}, renders the layout into a high-resolution PNG image~($\texttt{PNG}_\texttt{n}$). All generated PNG images are stored in a database to enable efficient retrieval during both fine-tuning and inference.
Each pair $\texttt{MD}_\texttt{n}$, $\texttt{PNG}_\texttt{n}$ is subsequently combined into a unified, high-level representation of the original GDSII layout~($\texttt{PNG}_\texttt{n}^{\texttt{*}}$). This representation encapsulates the complete set of extracted metadata together with a reference to the corresponding PNG image stored in the database. There exists a one-to-one correspondence between $\texttt{GDS}_\texttt{n}$ and $\texttt{PNG}_\texttt{n}^{\texttt{*}}$ for $\texttt{n} = 0, \ldots, \texttt{N}$, where $\texttt{N}$ denotes the total number of layouts.
In the second step, \textit{Q\&As Generation}, the pipeline processes the collection of $\texttt{PNG}_\texttt{n}^{\texttt{*}}$ representations to generate sets of task-specific questions and answers~(\texttt{Q\&As}). This step analyzes the metadata associated with each layout and constructs layout- and task-specific questions. Each \texttt{Q\&As} instance is associated with a specific $\texttt{PNG}_\texttt{n}$, includes a user query asking for a particular metadata attribute, and provides a corresponding ground-truth answer. Multiple \texttt{Q\&As} may reference the same $\texttt{PNG}_\texttt{n}$. Moreover, each task may have a different number of \texttt{Q\&As}, e.g., in Figure~\ref{sfig:dataset_creation}, Task A has $\texttt{L}_\texttt{A}$ \texttt{Q\&As}, Task B has $\texttt{L}_\texttt{B}$ \texttt{Q\&As} and so on.
Finally, the \textit{Splits Division} step organizes the generated \texttt{Q\&As} into task-specific sub-datasets and further partitions each sub-dataset into training, validation, and test splits.
All resulting splits are stored in a database and used in the subsequent fine-tuning and evaluation stages. Notably, all \texttt{Q\&As} generated from the same layout are assigned exclusively to a single split~(train, validation, or test). Consequently, no rendered image~(PNG) or underlying layout~(GDS) appears across different splits under distinct questions. 

\subsection{VLM Fine-tuning}
\label{ssec:vlm_finetuning}
Once the task-specific dataset has been constructed, the \textit{VLM Fine-tuning} stage is performed. A single-task training paradigm is adopted, in which a model is explicitly optimized to address one task, thereby promoting focused and task-specific learning. Figure~\ref{sfig:vlm_finetune} illustrates the fine-tuning process for a generic task. 
The fine-tuning pipeline is organized into three main steps: \emph{(i)} \textit{Get PNG}, \emph{(ii)} \textit{Pre-processing}, and \emph{(iii)} \textit{VLM Adaptation}. 
First, each training or validation \texttt{Q\&A} instance is de-coupled into its constituent questions~(\texttt{Q}) and answers~(\texttt{A}). While the textual components are directly forwarded to the \textit{Pre-processing} step, the associated image path is used in \textit{Get PNG} to retrieve the corresponding layout image~(\texttt{PNG}), which is then passed to the same stage.
The \textit{Pre-processing} step is further decomposed into two sub-steps: \emph{(i)} \textit{Image Manipulation} and \emph{(ii)} \textit{Chat Serialization}. In \textit{Image Manipulation}, each \texttt{PNG} is processed to match the patch-based input format expected by the vision encoder, ensuring that the spatial dimensions are compatible with its internal grid. In \textit{Chat Serialization}, each \texttt{Q\&A} exchange is converted into a single contiguous text sequence introducing special tokens that explicitly denote the user role for the question, the assistant role for the answer, message boundaries, and the end-of-sequence marker. The resulting structured transcript enables the model to learn to generate the response conditioned jointly on the visual input and the user query.
Finally, the \textit{VLM Adaptation} step applies a standard supervised fine-tuning procedure, where the model is trained to predict the ground-truth answer tokens given the prompt and image, using a masked next-token cross-entropy loss.
\begin{table*}[t]
	\centering
	\caption{VLMs considered in this work, compared by backbone, vision encoder, multimodal and instruction tuning, supported resolution, and spatial awareness. InternLM-4KHD and Qwen2.5-VL-7B are fine-tuned for the controlled backbone-transfer study.}
	\footnotesize
	\resizebox{\linewidth}{!}{
		\setlength{\tabcolsep}{6pt}
		\renewcommand{\arraystretch}{1.3}
		\begin{tabular}{lcccccccc}
			\toprule
			\rotatebox{0}{\textbf{Model}}
			& \rotatebox{0}{\textbf{Family}} 
			& \rotatebox{0}{\textbf{Params}} 
			& \rotatebox{0}{\textbf{LB}} 
			& \rotatebox{0}{\textbf{VE}} 
			& \rotatebox{0}{\textbf{MM Train.}} 
			& \rotatebox{0}{\textbf{Inst-tuned}} 
			& \rotatebox{0}{\textbf{Res.}} 
			& \rotatebox{0}{\textbf{Sp. Aw.}} \\ 
			
			\midrule
			
			GPT-5.2~\cite{gpt}
			& OpenAI 
			& --
			& Proprietary 
			& Proprietary 
			& \faCircle 
			& \faCircle 
			& \faCircle 
			& \faDotCircleO \\
			
			Gemini-3-Flash~\cite{gemini}
			& Google 
			& --
			& Proprietary 
			& Proprietary 
			& \faCircle 
			& \faCircle 
			& \faCircle 
			& \faDotCircleO \\
			
			\midrule
			
			GLM-4V-9B~\cite{ABB+2024} 
			& Zhipu AI 
			& $9$B
			& GLM 
			& ViT 
			& \faCircle 
			& \faCircle 
			& \faCircle 
			& \faDotCircleO \\
			
			InternLM-4KHD~\cite{XPY+2024} 
			& InternLM 
			& $7$B
			& InternLM 
			& ViT (4K) 
			& \faCircle 
			& \faCircle 
			& \faCircle 
			& \faDotCircleO \\
			
			InternVL2~\cite{CWC+2024} 
			& InternVL 
			& $8$B
			& InternLM 
			& Hybrid 
			& \faCircle 
			& \faCircle 
			& \faCircle 
			& \faCircle \\
			
			InternVL3.5~\cite{WZL+2025} 
			& InternVL 
			& $8$B
			& InternLM 
			& Hybrid 
			& \faCircle 
			& \faCircle 
			& \faCircle 
			& \faCircle \\
			
			LLaVA-Qwen-7B~\cite{HCQ+2023} 
			& LLaVA / Qwen 
			& $7$B
			& Qwen 
			& ViT 
			& \faCircle 
			& \faCircle 
			& \faDotCircleO 
			& \faCircleO \\
			
			Pixtral-12B~\cite{AAB+2024} 
			& Mistral 
			& $12$B
			& Mistral 
			& ViT 
			& \faCircle 
			& \faCircle 
			& \faCircle 
			& \faDotCircleO \\
			
			Qwen2.5-VL-7B~\cite{BCL+2025} 
			& Qwen 
			& $7$B
			& Qwen2.5 
			& ViT 
			& \faCircle 
			& \faCircle 
			& \faCircle 
			& \faDotCircleO \\
			
			Qwen3-VL-8B~\cite{SYR+2025} 
			& Qwen 
			& $8$B
			& Qwen3 
			& ViT 
			& \faCircle 
			& \faCircle 
			& \faCircle 
			& \faDotCircleO \\
			
			\bottomrule
		\end{tabular}
	}
	
	\vspace{1mm}
	\begin{minipage}{0.95\linewidth}
		\footnotesize
		\faCircle~Supported/High,
		\faDotCircleO~Partial/Medium,
		\faCircleO~Limited/absent. \\
		Language Backbone~(LB),
		Vision Encoder~(VE),
		Multimodal training~(MM Train.),
		Instruction tuned~(Instr-tuned),
		Resolution~(Res.),
		Spatial Awareness~(Sp. Aw.)
	\end{minipage}
	
	\label{tbl:vlm_models}
\end{table*}

\subsection{VLM Deployment}
\label{ssec:vlm_deployment}
Once a VLM has been fine-tuned for a given task, it is deployed on the corresponding test set. Figure~\ref{sfig:vlm_deployment} illustrates the \textit{VLM Deployment} stage. Test samples are processed individually. For each input instance, the procedure outputs a binary \texttt{Match}/\texttt{No Match} decision indicating whether the answer generated by the fine-tuned VLM is semantically consistent with the ground-truth response.
To maximize the efficiency of the proposed methodology and minimize human-in-the-loop, an LLM is employed to judge the semantic equivalence between the textual answer produced by our fine-tuned VLM and the ground truth. 
As in the \textit{VLM Fine-tuning} stage, each test \texttt{Q\&A} pair is first de-coupled. The associated layout image $\texttt{PNG}_\texttt{n}$ is retrieved, while the ground-truth answer $\texttt{A}\texttt{x}$ is forwarded to the judge and the question $\texttt{Q}_\texttt{x}$ is passed to the \textit{Pre-processing} step. Pre-processing mirrors the procedure used during fine-tuning, with the key difference that the ground-truth answer is not included in the chat serialization. The pre-processed question and image are then provided as input to the VLM, which generates a predicted answer $\texttt{A}_\texttt{x}^{\texttt{*}}$. Both the predicted answer and the ground-truth answer are submitted to the judge, which performs a semantic comparison to determine the final match outcome.

\section{Results}
\label{sec:results}
\begin{table}[t]
	\centering
	\caption{Analog circuits dataset composition and statistics. For each category, the table reports the circuit types, the number of layout variants, the average number of devices per variant, and the average per-device-type counts~(NMOS, PMOS, capacitors, and resistors).}
	\label{tbl:dataset_composition}
	\footnotesize
	\resizebox{\linewidth}{!}{%
		\begin{tabular}{llccccccc}
			\toprule
			\textbf{Category} & \textbf{Circuit Type} & \textbf{Variants}~(\#) & \textbf{Avg. Devices}~(\#) & \textbf{NMOS}~(\#) & \textbf{PMOS}~(\#) & \textbf{CAP}~(\#) & \textbf{RES}~(\#) & \textbf{Total Devices}~(\#) \\
			\midrule
			\multirow{5}{*}{\shortstack{Single\\Device}} & \quad Capacitor & $4\,997$ & $1$ & -- & -- & $1$ & -- & $4\,997$ \\
			& \quad NMOS Transistor & $5\,000$ & $1$ & $1$ & -- & -- & -- & $5\,000$ \\
			& \quad PMOS Transistor & $5\,000$ & $1$ & -- & $1$ & -- & -- & $5\,000$ \\
			& \quad Resistor & $5\,000$ & $1$ & -- & -- & -- & $1$ & $5\,000$ \\
			\cmidrule(lr){1-9}
			& \quad Subtotal & $19\,997$ & -- & -- & -- & -- & -- & $19\,997$ \\
			\midrule
			\multirow{7}{*}{\shortstack{Base\\Circuits}} & \quad Ahuja OTA & $995$ & $15$ & $10$ & $4$ & $1$ & -- & $14\,925$ \\
			& \quad Gate Driver & $1\,000$ & $10$ & $4$ & $4$ & -- & $2$ & $10\,000$ \\
			& \quad High-Pass Filter (HPF) & $962$ & $13$ & $5$ & $3$ & $3$ & $2$ & $12\,506$ \\
			& \quad Low-Dropout Regulator (LDO) & $989$ & $9$ & $3$ & $3$ & $1$ & $2$ & $8\,901$ \\
			& \quad Low-Pass Filter (LPF) & $971$ & $13$ & $5$ & $3$ & $3$ & $2$ & $12\,623$ \\
			& \quad Miller OTA & $977$ & $13$ & $5$ & $4$ & $2$ & $2$ & $12\,701$ \\
			\cmidrule(lr){1-9}
			& \quad Subtotal & $5\,894$ & -- & -- & -- & -- & -- & $71\,656$ \\
			\midrule
			\multirow{1}{*}{\shortstack{Mixed}} & \quad Mixed Topologies & $4\,140$ & $21.6$ & $9.7$ & $6.8$ & $2.3$ & $2.8$ & $89\,424$ \\
			\midrule
			& \textbf{Total Dataset} & \textbf{$30\,034$} & -- & -- & -- & -- & -- & \textbf{$181\,080$} \\
			\bottomrule
		\end{tabular}
	}
\end{table}
\begin{table*}[t]
	\centering
	\caption{Task descriptions, organized by complexity \emph{(i)} Easy, \emph{(ii)} Medium, and \emph{(iii)} Hard.}
	\label{tbl:training_tasks}
	\footnotesize
	\resizebox{\linewidth}{!}{%
		\begin{tabular}{lllr}
			\toprule
			\textbf{Complexity} & \textbf{Task ID} & \textbf{Task Description} & \textbf{Q\&As}~(\#) \\
			\midrule
			\multirow{1}{*}{\textbf{Easy}} 
			& A & Identification of single component devices (capacitors, resistors, NMOS, PMOS) & $19\,997$ \\
			\midrule
			\multirow{2}{*}{\textbf{Medium}} 
			& B & Identification of base circuit topologies (OTAs, filters, regulators, gate drivers) & $5\,894$ \\
			\cmidrule(lr){2-4}
			& C & Component counting and enumeration in base circuits & $27\,475$ \\
			\midrule
			\multirow{2}{*}{\textbf{Hard}} 
			& D & Component counting and enumeration in complex mixed circuits & $19\,848$ \\
			\cmidrule(lr){2-4}
			& E & Identification of base circuit topologies within complex mixed circuits & $4\,140$ \\
			\midrule
			\multicolumn{3}{l}{\textbf{Total}} & $77\,354$ \\
			\bottomrule
		\end{tabular}
	}
\end{table*}
\begin{figure}[t]
	\centering
	\begin{subfigure}[b]{0.23\columnwidth}
		\centering
		\includegraphics[width=0.6\linewidth]{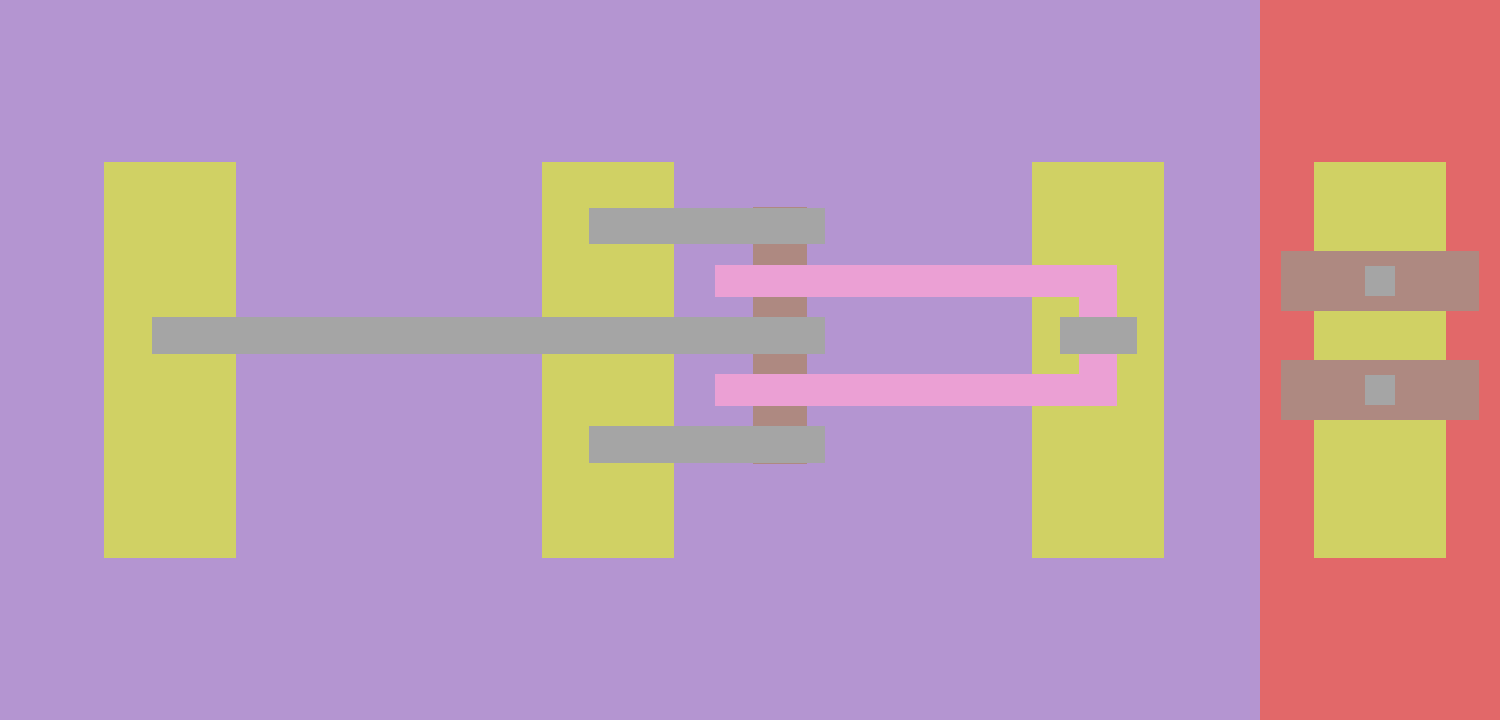}
		\caption{PMOS transistor.}
		\label{sfig:pmos}
	\end{subfigure}
	\begin{subfigure}[b]{0.23\columnwidth}
		\centering
		\includegraphics[width=0.6\linewidth]{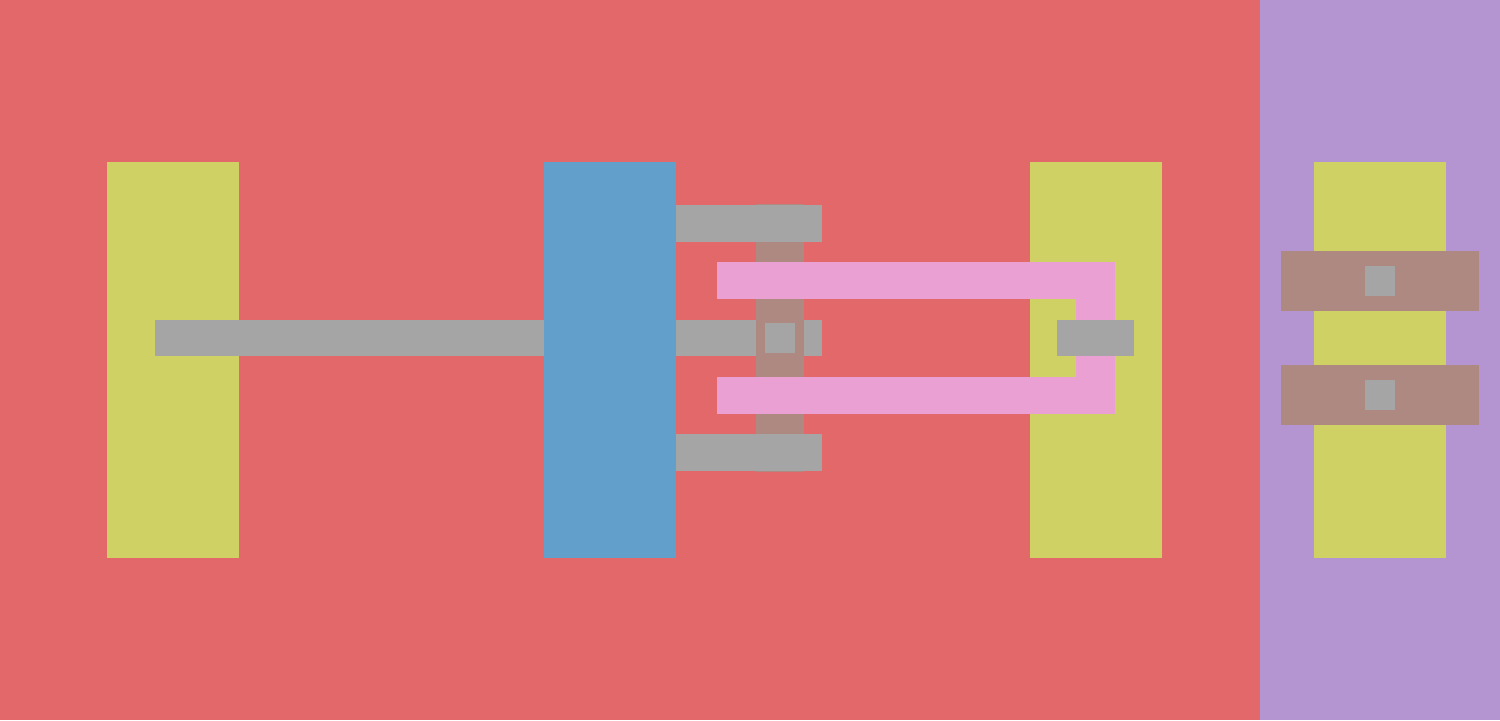}
		\caption{NMOS transistor.}
		\label{sfig:nmos}
	\end{subfigure}
	\begin{subfigure}[b]{0.23\columnwidth}
		\centering
		\rotatebox{0}{\includegraphics[width=\linewidth]{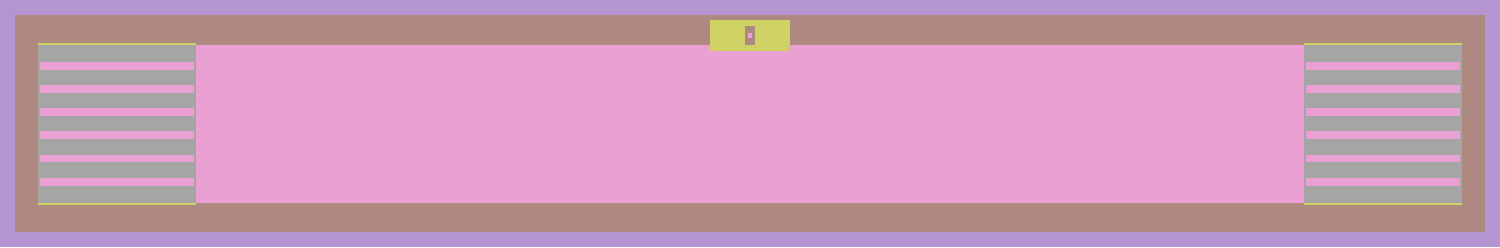}}
		\caption{Resistor.}
		\label{sfig:res}
	\end{subfigure}
	\begin{subfigure}[b]{0.23\columnwidth}
		\centering
		\includegraphics[width=0.4\linewidth]{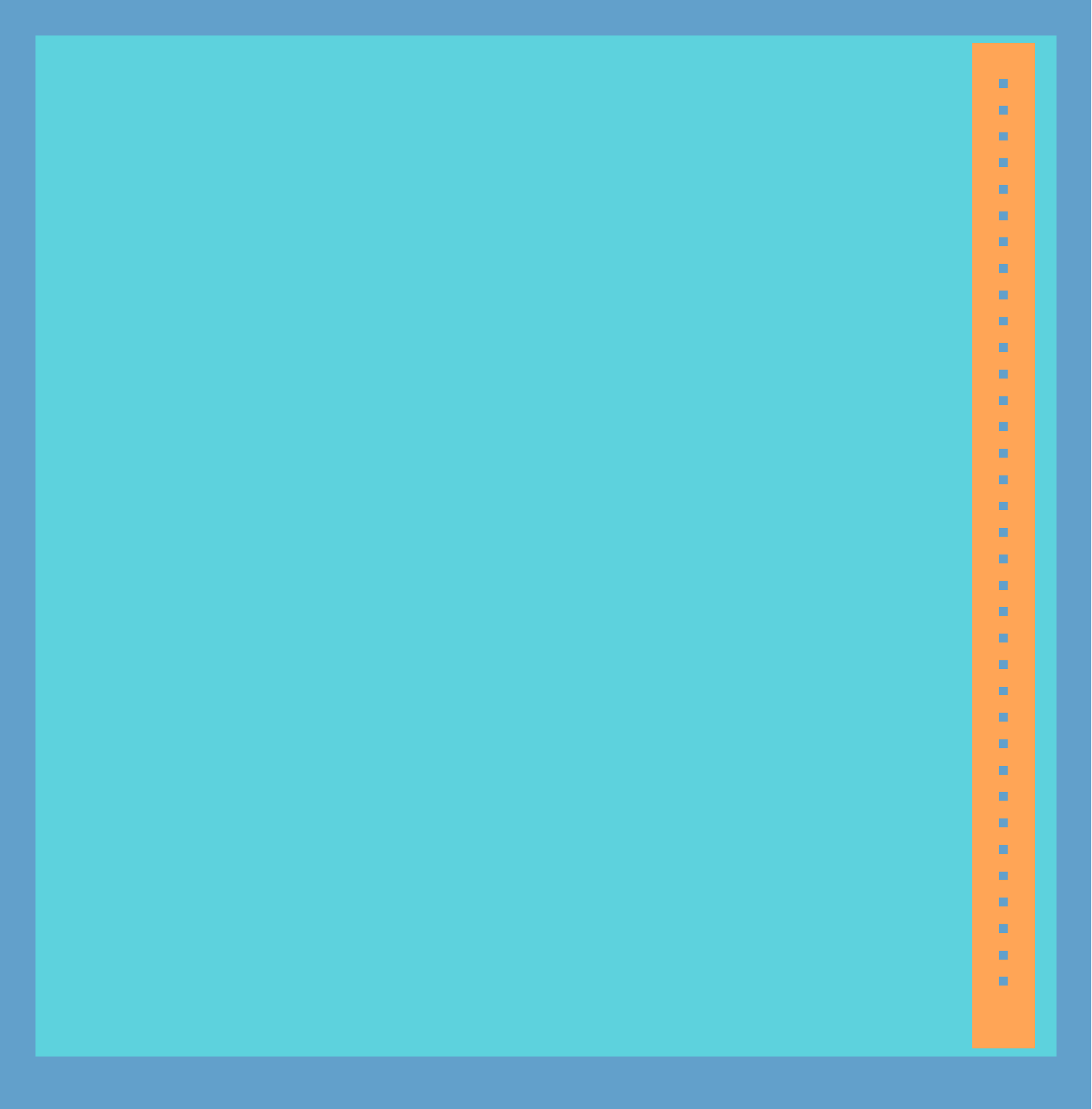}
		\caption{Capacitor.}
		\label{sfig:cap}
	\end{subfigure}
	\caption{Examples of GDSII view of single-component devices layouts.}
	\label{fig:single_components_examples}
\end{figure}
This section is divided into two parts. Section~\ref{ssec:experimental_results} describes the experimental setup, whereas Section~\ref{ssec:inference_results} reports the performance, transfer, and evaluation-robustness analyses.

\subsection{Experimental Setup}
\label{ssec:experimental_results}
\noindent\textbf{Hardware setup -} The experimental evaluation was conducted on a workstation running Ubuntu $24.04$ with $512$ GB of RAM and $4\times$ NVIDIA A100~($40$ GB) configuration. The GPU was employed to carry out both fine-tuning and inference of VLMs. 

\noindent\textbf{Software setup -} The experiments were carried out using \texttt{pytorch}, Hugging Face's \texttt{transformers}, and \texttt{peft} for LoRA~\cite{HWA+2022}.
Qwen3-32B~\cite{AAB+2025} was used as LLM-as-a-judge to carry out semantic comparison between ground truth answers and VLM-generated outputs.
Moreover, image resizing and padding are performed, during \textit{Image Manipulation} in accordance with the requirements specified by InternLM-4KHD.
Further details on InternLM-4KHD parameters configuration are discussed in Appendix~\ref{ssec:ft_config}.
Train, validation, and test splits were constructed by building balanced sets across all tasks and circuit families. This splitting process was carried out at the level of task-specific sub-datasets, organizing the generated Q\&As to maintain a representative distribution of the diverse circuit categories within each split. 
Table~\ref{tbl:vlm_models} reports a comparative overview of the evaluated VLMs.

\noindent\textbf{Dataset composition -} Table~\ref{tbl:dataset_composition} summarizes the composition of the analog circuits dataset used in this work, detailing circuit categories, structural diversity, and device-level statistics. The dataset is organized into three main groups: \emph{(i)} \texttt{Single Component}, \emph{(ii)} \texttt{Base Circuit}, and \emph{(iii)} \texttt{Mixed}, capturing a wide spectrum of design complexity.
The \texttt{Single Component} group comprises $20\,000$ variants evenly distributed across capacitors, NMOS transistors, PMOS transistors, and resistors. Each variant contains a single component, providing a controlled setting for low-level visual and semantic recognition tasks and serving as a foundation for basic component understanding. Figure~\ref{fig:single_components_examples} reports a layout example for a PMOS transistor~(Figure~\ref{sfig:pmos}), 
NMOS transistor~(Figure~\ref{sfig:nmos}), resistor~(Figure~\ref{sfig:res}), and capacitor~(Figure~\ref{sfig:cap}).
The \texttt{Base Circuit} group includes $5\,894$ variants spanning six commonly used analog building blocks, such as operational transconductance amplifiers~(OTAs), filters, gate drivers, and regulators. These circuits exhibit moderate structural complexity, with an average of $9$ to $15$ components per layout. 
The \texttt{Mixed } group consists of $4\,140$ variants that combine multiple base topologies into larger layouts, significantly increasing structural diversity and components count. These circuits average $21.6$ components per variant and include heterogeneous mixtures of devices, posing more challenging recognition and reasoning demands.
This hierarchical composition enables systematic evaluation across increasing levels of circuit complexity and supports the analysis of VLM capabilities ranging from single-device identification to reasoning over complex, mixed analog layouts.
All layouts are implemented in the SkyWater 130nm PDK and satisfy both design rule check~(DRC) 
and layout-vs-schematic~(LVS) requirements. Appendix~\ref{ssec:dataset_structure_statistics} provides further details on dataset structure and statistics, while Appendix~\ref{ssec:released_data} shows a few layouts contained in the dataset.

\noindent\textbf{Tasks specification -} Table~\ref{tbl:training_tasks} presents an overview of the tasks defined in this work, organized by increasing levels of complexity. Notably, the level of complexity for each task has been experimentally selected according to the results obtained with state-of-the-art models which results have been discussed in Section~\ref{sec:introduction}~(see Figure~\ref{fig:motivational_example}).
Each task is characterized by its semantic objective, the structural difficulty of the underlying circuit layouts, and the corresponding number of \texttt{Q\&As}.
The \texttt{Easy} category includes \texttt{Task~A}, which focuses on the identification of single-component devices such as capacitors, resistors, NMOS, and PMOS transistors. With $20\,000$ \texttt{Q\&As}, this task evaluates the model’s ability to recognize basic circuit primitives and serves as an entry point for learning low-level visual features and device semantics.
The \texttt{Medium} complexity category comprises two tasks. \texttt{Task~B} addresses the identification of base circuit topologies, including OTAs, filters, regulators, and gate drivers, requiring the model to reason over structured combinations of multiple components. \texttt{Task~C} further increases the difficulty by asking the model to count and enumerate components within these base circuits, emphasizing quantitative reasoning and spatial awareness. Together, these tasks account for $33\,369$ \texttt{Q\&As} and assess the transition from component-level recognition to circuit-level understanding.
The \texttt{Hard} category targets complex mixed circuits that integrate multiple base topologies within a single layout. \texttt{Task~D} requires component counting and enumeration in these heterogeneous layouts, while \texttt{Task~E} focuses on identifying base circuit topologies embedded within complex designs. These tasks are the most challenging, as they demand simultaneous reasoning over dense layouts, overlapping functional blocks, and diverse device types. The datasets presents $23\,988$ \texttt{Q\&As} belonging to the \texttt{Hard} category.
Overall, THEIA spans a total of $77\,354$ \texttt{Q\&As} across five tasks and three complexity tiers, reflecting a structured task hierarchy that progressively increases visual and semantic complexity. This organization enables a systematic evaluation of VLMs capabilities from elementary component recognition to the analysis of complex analog circuit layouts.
The proposed tasks are designed to approximate recurring sub-problems in analog back-end workflows. In particular, tasks such as component identification, counting, and topology recognition are implicit steps in layout debugging and verification as well as in signoff analysis, where designers manually inspect layout structures to diagnose violations, mismatches, or unexpected behavior. Our formulation abstracts these operations into controlled evaluation tasks, enabling systematic benchmarking of visual reasoning capabilities. 
%
\begin{table*}[t]
	\centering
	\caption{Accuracy on the held-out THEIA test splits. The Qwen2.5-VL-7B transfer uses the same splits, prompts, answer format, and deterministic decoding with and without task-specific fine-tuning.}
	\label{tbl:perf_comparison}
	\scriptsize
	\resizebox{0.8\textwidth}{!}{%
		\begin{tabular}{llcrrrrrr}
			\toprule
			\multirow{2}{*}{\textbf{Complexity}} & \multirow{2}{*}{\textbf{Task}} & \multirow{2}{*}{\textbf{Count}~(\#)} & \multicolumn{2}{c}{\textbf{InternLM FT}} & \multirow{2}{*}{\shortstack{\textbf{InternLM}\\\textbf{no FT}}} & \multirow{2}{*}{\textbf{GPT-5.2}} & \multicolumn{2}{c}{\textbf{Qwen2.5-VL-7B}} \\
			\cmidrule(lr){4-5}\cmidrule(lr){8-9}
			& & & \textbf{Pass@1} & \textbf{Pass@5} & & & \textbf{no FT} & \textbf{FT} \\
			\midrule
			\multirow{1}{*}{\textbf{Easy}} & A & $1\,001$ & \textbf{100\%} & \textbf{100\%} & $41\%$ & $100\%$ & $19\%$ & \textbf{100\%} \\
			\midrule
			\multirow{2}{*}{\textbf{Medium}} & B & $300$ & \textbf{84\%} & \textbf{92\%} & $16\%$ & $83\%$ & $17\%$ & \textbf{85\%} \\
			\cmidrule(lr){2-9}
			& C & $1\,399$ & \textbf{83\%} & \textbf{94\%} & $18\%$ & $20\%$ & $23\%$ & \textbf{93\%} \\
			\midrule
			\multirow{2}{*}{\textbf{Hard}} & D & $1\,003$ & \textbf{63\%} & \textbf{87\%} & $24\%$ & $26\%$ & $24\%$ & \textbf{89\%} \\
			\cmidrule(lr){2-9}
			& E & $207$ & \textbf{81\%} & \textbf{89\%} & $7\%$ & $10\%$ & $20\%$ & \textbf{91\%} \\
			\bottomrule
		\end{tabular}}
\end{table*}
\begin{table}[t]
	\centering
	\caption{Cross-generator evaluation of InternLM-4KHD fine-tuned on THEIA and evaluated on held-out ALIGN~\cite{KMS+2019} layouts.}
	\label{tbl:cross_generator}
	\scriptsize
	\resizebox{0.7\linewidth}{!}{%
		\begin{tabular}{lccc}
			\toprule
			\textbf{Task} & \textbf{THEIA test} & \textbf{ALIGN test} & \textbf{$\Delta$} \\
			\midrule
			B: topology identification & $84.0\%$ ($252/300$) & $89.5\%$ ($34/38$) & $+5.5$ pp \\
			C: component counting & $83.0\%$ ($1\,161/1\,399$) & $73.7\%$ ($84/114$) & $-9.3$ pp \\
			\bottomrule
		\end{tabular}}
\end{table}
\begin{table}[t]
	\centering
	\caption{Deterministic validation metrics for fine-tuned InternLM-4KHD on held-out THEIA samples. LLM-judge Pass@1 is shown for comparison.}
	\label{tbl:deterministic_validation}
	\scriptsize
	\resizebox{0.6\linewidth}{!}{%
		\begin{tabular}{lccccc}
			\toprule
			\textbf{Task} & \textbf{Judge Pass@1} & \textbf{EM} & \textbf{NSM} & \textbf{Count Acc.} & \textbf{Topology-set F1} \\
			\midrule
			A & $100\%$ & $100\%$ & $100\%$ & -- & -- \\
			B & $84\%$ & $84\%$ & $84\%$ & -- & -- \\
			C & $83\%$ & $83\%$ & $83\%$ & $83\%$ & -- \\
			D & $63\%$ & $63\%$ & $63\%$ & $63\%$ & -- \\
			E & $81\%$ & $81\%$ & $81\%$ & -- & $89\%$ \\
			\bottomrule
	\end{tabular}}
\end{table}
\begin{table}[t]
	\centering
	\caption{Agreement of Qwen3-32B LLM judge with deterministic evaluation on held-out samples.}
	\label{tbl:judge_agreement}
	\scriptsize
	\resizebox{0.6\linewidth}{!}{%
		\begin{tabular}{lrrrrr}
			\toprule
			\textbf{Task} & \textbf{Count} & \textbf{Cohen's $\kappa$} & \textbf{Accuracy} & \textbf{Precision} & \textbf{Recall / F1} \\
			\midrule
			A & $1\,001$ & $1.0$ & $100\%$ & $100\%$ & $100\%$ / $100\%$ \\
			B & $300$ & $1.0$ & $100\%$ & $100\%$ & $100\%$ / $100\%$ \\
			C & $1\,399$ & $1.0$ & $100\%$ & $100\%$ & $100\%$ / $100\%$ \\
			D & $1\,003$ & $1.0$ & $100\%$ & $100\%$ & $100\%$ / $100\%$ \\
			E & $207$ & $1.0$ & $100\%$ & $100\%$ & $100\%$ / $100\%$ \\
			\bottomrule
	\end{tabular}}
\end{table}
\begin{figure}[t]
	\centering
	\includegraphics[width=0.45\linewidth]{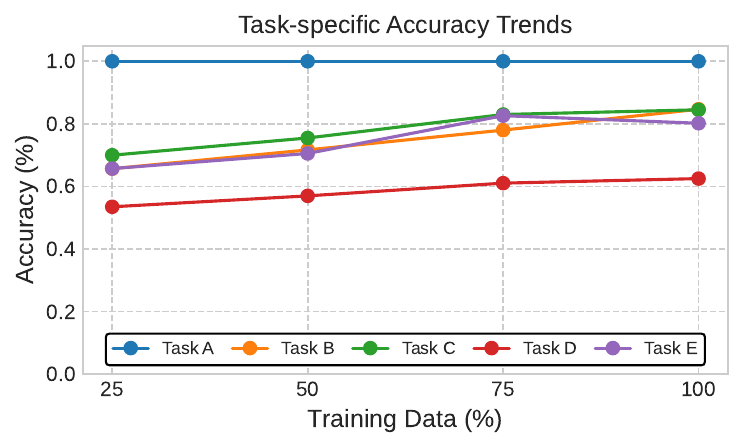}
	\caption{Accuracy trends as the amount of task-specific fine-tuning data is progressively increased.}
	\label{fig:ablation}
\end{figure}
\begin{table}[t]
	\centering
	\caption{Fine-tuned InternLM-4KHD accuracy for multiple-choice and open-ended forms of \texttt{Tasks~A-D} on held-out layouts.}
	\label{tbl:open_ended}
	\scriptsize
	\begin{tabular}{ccc}
		\toprule
		\textbf{Task} & \textbf{Multiple choice} & \textbf{Open ended} \\
		\midrule
		A & $100\%$ & $100\%$ \\
		B & $84\%$ & $83\%$ \\
		C & $83\%$ & $99\%$ \\
		D & $63\%$ & $75\%$ \\
		\bottomrule
	\end{tabular}
\end{table}

\subsection{Experimental Results}
\label{ssec:inference_results}
Table~\ref{tbl:perf_comparison} provides a detailed performance breakdown tasks, grouped by complexity level~(\texttt{Easy}, \texttt{Medium}, and \texttt{Hard}) and task identifier. For every configuration, it reports the number of held-out samples~(\texttt{Count}) and the \texttt{Accuracy}~(measuring the fraction of correctly predicted answers) of task-specific fine-tuned InternLM-4KHD, its zero-shot counterpart, GPT-5.2, and, Qwen2.5-VL-7B before and after task-specific fine-tuning. 
InternLM-4KHD and Qwen2.5-VL-7B have been selected to demonstrate the effectiveness of the proposed fine-tuning methodology. 
GPT5.2, instead, has been selected as the best commercial, large, state-of-the-art solution according to the results in Figure~\ref{fig:motivational_example}. InternLM-4KHD obtains $100\%$ on \texttt{Task~A}, $84\%$ and $83\%$ on \texttt{Task~B} and \texttt{Task~C}, and $63\%$ and $81\%$ on \texttt{Task~D} and \texttt{Task~E}, respectively. The zero-shot InternLM-4KHD and GPT-5.2 results are substantially lower on the \texttt{Medium} and \texttt{Hard} tasks, although GPT-5.2 remains competitive on the simplest task. GPT-5.2 performance degrades sharply as task structure and ambiguity increase, underscoring the limitations of general-purpose VLMs when applied on niche domain tasks without adaptation.
The Qwen2.5-VL-7B transfer improves all five tasks by $65$-$81$ percentage points, from $19\%$-$24\%$ zero-shot accuracy to $85\%$-$100\%$ after fine-tuning. Because the splits, prompts, answer format, and deterministic decoding are identical before and after fine-tuning, these gains isolate the effect of adaptation and demonstrate transfer across backbones. Overall, the Table~\ref{tbl:perf_comparison} quantitatively demonstrates that task-specific fine-tuning is essential to enable reliable understanding over analog layout data.
The evaluation isolates domain adaptation by comparing each backbone before and after task-specific fine-tuning under a controlled protocol. It is not intended as a general ranking of backbone architectures. For closed-set \texttt{Task~A} and \texttt{Task~B}, CNN baselines from Inspector~\cite{CCP+2026} achieve $97\%$ and $91\%$, respectively. These classifiers provide a task-specific closed-set reference, whereas THEIA evaluates a single VLM across open-ended component, topology, and counting queries.
To assess generator transfer, Table~\ref{tbl:cross_generator} evaluates InternLM-4KHD, originally fine-tuned on THEIA, on held-out ALIGN layouts~\cite{KMS+2019} from the same PDK. The result is encouraging for topology identification~($89.5\%$, $34/38$) but component counting decreases to $73.7\%$ ($84/114$).

\subsubsection{Evaluation Robustness}
\label{ssec:evaluation_robustness}
Table~\ref{tbl:deterministic_validation} reports deterministic metrics for the fine-tuned InternLM-4KHD on the held-out THEIA samples. Exact match and normalized string match coincide with LLM-judge Pass@1 for every applicable task. The task-specific count accuracy also coincides for \texttt{Tasks~C} and \texttt{D}, while \texttt{Task~E} achieves $89\%$ topology-set F1. Table~\ref{tbl:judge_agreement} further reports agreement between the Qwen3-32B LLM judge and deterministic evaluation.
The multiple-choice protocol does not use a fixed number or position of answer options. For counting \texttt{Tasks~C} and \texttt{D}, distractors are formed from nearby integer counts before valid random counts are used, and the candidate order is shuffled. Appendix~\ref{sec:additional_evaluation} reports the resulting option-count distributions. \texttt{Task~B} is class-balanced across its six topology families, which each occupy $16.3\%$-$17.1\%$ of every split. Its $16.7\%$ majority-class baseline is substantially below the $84\%$ fine-tuned-VLM accuracy.
Open-ended variants of \texttt{Tasks~A-D} are evaluated on the same held-out layouts to test whether performance depends on the answer choices. Table~\ref{tbl:open_ended} shows unchanged \texttt{Task-A} accuracy, a one-percentage-point reduction for \texttt{Task~B}, and improved accuracy on the counting tasks.
Figure~\ref{fig:ablation} reports accuracy trends as the amount of task-specific fine-tuning data is progressively increased~($25$\%, $50$\%, $75$\%, $100$\%) while keeping the evaluation protocol and the held-out test set fixed.
\texttt{Task~A} saturates immediately, indicating that single-device recognition is learned with minimal data. \texttt{Task~B} and \texttt{Task~C} shows a monotonic improvement as training data increases, demonstrating that base-circuit identification benefits substantially from additional examples. \texttt{Task~D} and \texttt{E} accuracies also trend upward with more data, albeit with less smooth curves due to higher task complexity.

\section{Conclusions}
\label{sec:conclusions}
This work introduced THEIA, a multimodal dataset and benchmark for analog layout understanding, in which verified GDSII layouts are paired with visual question-answering tasks that span component identification, topology identification, device counting, and mixed-topology reasoning. The layouts are generated from human-designed circuit templates and validated through DRC and LVS checks, providing a realistic basis for evaluating layout semantics rather than raw geometry alone. Across the five tasks, the results show that unadapted general-purpose VLMs struggle as task structure and ambiguity increase, whereas task-specific fine-tuning substantially improves performance. Under a controlled protocol, fine-tuning Qwen2.5-VL-7B improves accuracy by $65$-$81$ percentage points and confirms that the adaptation result transfers across backbones. The released data, fine-tuning pipeline, and evaluation implementations support reproducible investigation of multimodal analog-layout reasoning and future interactive CAD tools.

\section{Acknowledgments}
\label{sec:acknowledgments}
This work was supported by Italian Ministero dell'Universita e della Ricerca (MUR)'s 
Fondo Italiano Scienze Applicate (FISA) under Grant no. FISA-2024-00333.


\section{Limitations}
\label{sec:limitations}
Despite its contributions, this work has limitations. First, although THEIA spans a range of representative analog components and topologies, from basic building blocks to more complex systems, it does not capture the full diversity and variability of analog designs. Second, the experimental evaluation focuses on a single technology node~(SkyWater 130 nm PDK) and does not explicitly assess cross-node generalization. This choice reflects common industry practice, where technology migration is costly and infrequent. Broader cross-node validation is left to future work.

\bibliographystyle{unsrt}
\bibliography{bibliography}


\appendix
\newpage
\section{Appendix}
\label{sec:a_appendix}
\subsection{Fine-tuning \& Deployment Configuration}
\label{ssec:ft_config}
Table~\ref{tbl:ft_config_experiment} reports the parameters configuration used to fine-tune and evaluate InternLM-XComposer2-4KHD~\cite{XPY+2024}.

\begin{table}[H]
	\captionof{table}{Fine-tuning and inference parameters.}
	\label{tbl:ft_config_experiment}
	\vspace{2pt}
	\scriptsize
	\setlength{\tabcolsep}{3pt}
	\renewcommand{\arraystretch}{0.88}
	\begin{tabular}{@{}p{0.25\textwidth} p{0.20\textwidth} p{0.53\textwidth}@{}}
		\toprule
		\textbf{Parameter} & \textbf{Value / Default} & \textbf{Meaning / Notes} \\
		\midrule
		\multicolumn{3}{@{}l}{\textbf{A. Base model and task formulation}}\\[1pt]
		\texttt{base\_model} & \texttt{internlm/internlm- xcomposer2-4khd-7b} & Pretrained VLM used as initialization (loaded with custom model code). \\
		\texttt{task\_type} & causal language modeling & Supervised instruction tuning in a chat-style transcript format. \\
		\texttt{data\_format} & single-turn, single-image & Each sample contains one image and one question-answer pair (serialized with role and boundary tokens). \\[2pt]
		
		\multicolumn{3}{@{}l}{\textbf{B. Optimization and training schedule}}\\[1pt]
		\texttt{num\_train\_epochs} & $1$ & Number of training epochs. \\
		\texttt{per\_device\_train\_batch\_size} & $1$ & Trainer micro-batch size. \\
		\texttt{gradient\_accumulation\_steps} & $8$ & Accumulate gradients for $8$ steps before each optimizer update. \\
		\texttt{effective\_batch\_size} & $1 \times 8$ (per device) & Effective batch per device is \texttt{per\_device\_train\_batch\_size} $\times$ \texttt{gradient\_accumulation\_steps} (global batch also depends on number of devices). \\
		\texttt{optimizer} & AdamW & Optimizer used by the Hugging Face Trainer. \\
		\texttt{learning\_rate} & $5\times 10^{-5}$ & Base learning rate. \\
		\texttt{weight\_decay} & $0.1$ & Weight decay regularization. \\
		\texttt{warmup\_ratio} & $0.01$ & Fraction of total steps used for LR warmup. \\
		\texttt{lr\_scheduler\_type} & cosine & Learning rate schedule after warmup. \\[2pt]
		
		\multicolumn{3}{@{}l}{\textbf{C. Precision and memory-related training options (algorithmic)}}\\[1pt]
		\texttt{mixed\_precision} & bf16 & Mixed-precision training is used for efficiency. \\
		\texttt{gradient\_checkpointing} & True & Activation checkpointing to reduce memory usage (at the cost of extra compute). \\[2pt]
		
		\multicolumn{3}{@{}l}{\textbf{D. Sequence and vision preprocessing}}\\[1pt]
		\texttt{max\_length} & $8192$ & Maximum token context length used during training. \\
		\texttt{img\_preprocessing} & resize + pad + normalize & Images are resized (aspect ratio preserved), padded for alignment, converted to tensors, and normalized. \\
		\texttt{img\_size} & $490$ & Image size setting used by the pipeline. \\
		\texttt{hd\_num} & $6$ & Controls the high-definition resize budget in the image transform. \\[2pt]
		
		\multicolumn{3}{@{}l}{\textbf{E. LoRA (parameter-efficient fine-tuning)}}\\[1pt]
		\texttt{use\_lora} & True & LoRA adapters are trained while the base model weights are kept fixed. \\
		\texttt{lora\_r} & $64$ & Adapter rank $r$ (capacity). \\
		\texttt{lora\_alpha} & $64$ & Scaling $\alpha$; effective scale is $\alpha/r$. \\
		\texttt{lora\_dropout} & $0.05$ & Dropout applied on the LoRA branch (regularization). \\
		\texttt{lora\_target\_modules} & attention.wqkv, attention.wo, feed\_forward.w1/w2/w3 & Transformer submodules to which LoRA adapters are applied. \\
		\texttt{lora\_bias} & none & Bias handling for LoRA (none means biases are not trained/saved). \\[2pt]
		
		

        \multicolumn{3}{@{}l}{\textbf{F. Inference (generation) parameters}}\\[1pt]
        \texttt{do\_sample} & False & Deterministic decoding (greedy); no random sampling. \\
        \texttt{num\_beams} & 1 & Number of beams for beam search (1 = greedy decoding). \\
        \texttt{temperature} & 1.0 & Softmax temperature (only effective when \texttt{do\_sample}=True). \\
        \texttt{top\_p} & 0.8 & Nucleus sampling threshold (only effective when \texttt{do\_sample}=True). \\
        \texttt{max\_new\_tokens} & 1024 & Maximum number of tokens to generate (e.g., 1 for single-choice tasks). \\
        \texttt{use\_cache} & False & KV-cache disabled during generation. \\
        \texttt{hd\_num} (inference) & 6 & Same HD budget used at training; ensures consistent image preprocessing. \\
        \texttt{max\_length} (inference) & 8192 & Maximum context length for generation. \\[2pt]
		\bottomrule
	\end{tabular}
\end{table}
%
%
%
%
\subsection{Dataset Structure and Statistics}
\label{ssec:dataset_structure_statistics}
\begin{figure}[H]
	\centering
	\includegraphics[width=0.7\linewidth]{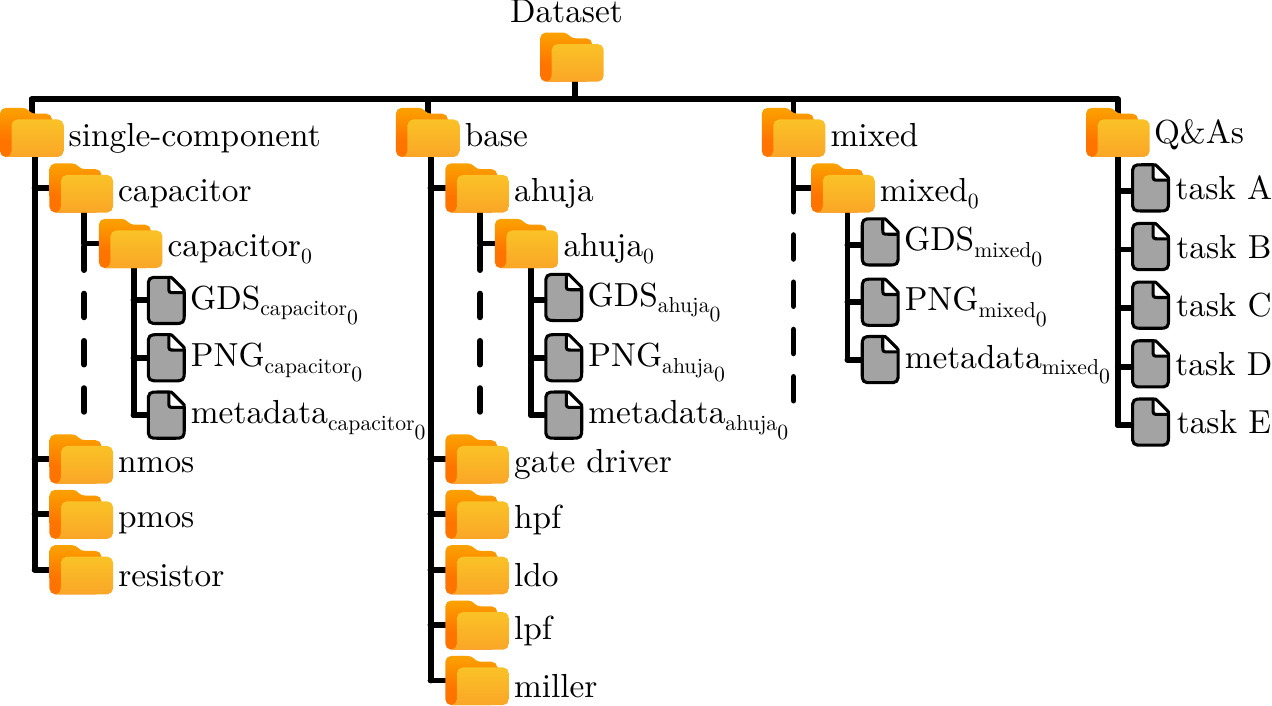}
	\caption{Dataset structure.}
	\label{fig:dataset_structure}
\end{figure}

Figure~\ref{fig:dataset_structure} illustrates the organization of the dataset. It is structured into four main directories: \emph{(i)} \texttt{single-component}, which includes layouts of individual devices, \emph{(ii)} \texttt{base}, containing base circuit topologies, \emph{(iii)} \texttt{mixed}, comprising complex mixed circuit layouts, and \emph{(iv)} \texttt{Q\&As}, which stores the task-specific question-answer sets. Within each circuit category, layouts are further organized by variant, with each variant directory containing the corresponding \texttt{GDS} file, the rendered \texttt{PNG} image, and an associated \texttt{metadata} file.
Table~\ref{tbl:dataset_details_appendix} reports the distribution of question-answer pairs across training, validation, and test splits, organized by task and complexity level. The dataset comprises $77\,354$ Q\&As in total, with $19\,997$ easy, $33\,369$ medium, and $23\,988$ hard samples. For each task, the table also details the circuit type composition, highlighting balanced distributions for simple tasks and exclusively mixed topologies for hard tasks.
Moreover, Table~\ref{tbl:comprehensive_stats} summarizes key geometric and device-level statistics of the dataset across layout categories and topologies. Layout area ($\mu m^2$) captures the spatial footprint of each design, while polygon count and layer count provide proxies for geometric and process complexity. The aspect ratio characterizes global layout shape. Device-level parameters include transistor width and length~($nm$), and number of fingers. Capacitor and resistor dimensions~($\mu m$) describe the physical realization of passive components, with capacitor area~($\mu m^2$) reported as a dataset-wide range to indicate capacitance variability. Finally, routing density is $\rho_{\mathrm{route}}=\sum_{\ell\in\mathcal{M}} A_{\ell}/A_{\mathrm{bbox}}$, where $A_{\ell}$ is the polygon area on routed-metal layer $\ell$, $\mathcal{M}$ is the set of SkyWater routed-metal layers, and $A_{\mathrm{bbox}}$ is the layout bounding-box area. Collectively, these metrics capture both structural and parametric diversity, supporting realistic and challenging reasoning tasks.
\begin{table}[H]
	\centering
	\caption{Subsets splits cardinality and balance analysis.}
	\label{tbl:dataset_details_appendix}
	\scriptsize
		\begin{tabular}{llrrrrl}
			\toprule
			\textbf{Complexity} & \textbf{Task} & \textbf{Train}~(\#) & \textbf{Validation}~(\#) & \textbf{Test}~(\#) & \textbf{Total}~(\#) & \textbf{Circuit Type Distribution} \\
			\midrule
			\multirow{4}{*}{\textbf{Easy}}
			& \multirow{4}{*}{\textbf{A}} 
			& \multirow{4}{*}{$17\,997$} 
			& \multirow{4}{*}{$999$} 
			& \multirow{4}{*}{$1\,001$} 
			& \multirow{4}{*}{$19\,997$} 
			& NMOS: $5\,000$ ($25.0\%$) \\
			& & & & & & PMOS: $5\,000$ ($25.0\%$) \\
			& & & & & & RES: $5\,000$ ($25.0\%$) \\
			& & & & & & CAP: $4\,99$7 ($25.0\%$) \\
			\midrule
			\multirow{12}{*}{\textbf{Medium}}
			& \multirow{6}{*}{\textbf{B}} 
			& \multirow{6}{*}{$5\,302$} 
			& \multirow{6}{*}{$292$} 
			& \multirow{6}{*}{$300$} 
			& \multirow{6}{*}{$5\,894$} 
			& Gate Driver: $1\,000$ ($17.0\%$) \\
			& & & & & & Ahuja OTA: $995$ ($16.9\%$) \\
			& & & & & & LDO: $989$ ($16.8\%$) \\
			& & & & & & Miller OTA: $977$ ($16.6\%$) \\
			& & & & & & LPF: $971$ ($16.5\%$) \\
			& & & & & & HPF: $962$ ($16.3\%$) \\
			\cmidrule(lr){2-7}
			& \multirow{6}{*}{\textbf{C}} 
			& \multirow{6}{*}{$24\,715$} 
			& \multirow{6}{*}{$1\,361$} 
			& \multirow{6}{*}{$1\,399$} 
			& \multirow{6}{*}{$27\,475$} 
			& LDO: $4\,945$ ($18.0\%$) \\
			& & & & & & Miller OTA: $4\,885$ ($17.8\%$) \\
			& & & & & & LPF: $4\,855$ ($17.7\%$) \\
			& & & & & & HPF: $4\,810$ ($17.5\%$) \\
			& & & & & & Gate Driver: $4\,000$ ($14.6\%$) \\
			& & & & & & Ahuja OTA: $3\,980$ ($14.5\%$) \\
			\midrule
			\multirow{2}{*}{\textbf{Hard}}
			& \textbf{D} 
			& $17\,842$
			& $1\,003$
			& $1\,003$
			& $19\,848$
			& Mixed circuits: $19\,848$ ($100\%$) \\
			\cmidrule(lr){2-7}
			& \textbf{E} 
			& $3\,726$
			& $207$
			& $207$
			& $4\,140$
			& Mixed circuits: $4\,140$ ($100\%$) \\
			\midrule
			\multicolumn{2}{l}{\textbf{Total Q\&A Pairs}~(\#)} & $69\,585$ & $3\,910$ & $3\,862$ & $77\,354$ & \\
			\bottomrule
		\end{tabular}
\end{table}
\begin{table}[H]
	\centering
	\caption{Dataset statistics across groups and circuit topologies.}
	\label{tbl:comprehensive_stats}
	\scriptsize
	\resizebox{\linewidth}{!}{%
	\begin{tabular}{l ccc ccc ccc ccc ccc}
		\toprule
		& \multicolumn{3}{c}{Layout area~($\mu m^2$)} 
		& \multicolumn{3}{c}{Polygon count~(\#)} 
		& \multicolumn{3}{c}{Layers~(\#)} 
		& \multicolumn{3}{c}{Aspect ratio~(-)} 
		& \multicolumn{3}{c}{Routing density~(-)} \\
		\cmidrule(lr){2-4} \cmidrule(lr){5-7} \cmidrule(lr){8-10} \cmidrule(lr){11-13} \cmidrule(lr){14-16}
		Topology & p25 & med & p75 & p25 & med & p75 & p25 & med & p75 & p25 & med & p75 & p25 & med & p75 \\
		\midrule
		\multicolumn{16}{l}{Dataset groups} \\
		Single Devices & $94$ & $125$ & $663$ & $43$ & $51$ & $228$ & $8$ & $9$ & $10$ & $0.6$ & $0.6$ & $2.1$ & $5.5$ & $5.9$ & $6.5$ \\
		Base Circuits      & $2\,965$ & $6\,149$ & $9\,828$ & $4\,073$ & $4\,674$ & $5\,471$ & $13$ & $14$ & $14$ & $1.0$ & $1.5$ & $2.1$ & $3.7$ & $4.0$ & $4.6$ \\
		Mixed      & $6\,209$ & $10\,545$ & $14\,887$ & $9\,407$ & $13\,109$ & $17\,365$ & $14$ & $14$ & $15$ & $0.7$ & $1.3$ & $2.0$ & $3.5$ & $3.8$ & $4.3$ \\
		\midrule
		\multicolumn{16}{l}{Single Devices} \\
		Capacitor        & $607$ & $1\,907$ & $4\,075$ & $28$ & $51$ & $75$ & $4$ & $4$ & $4$ & $1.0$ & $1.0$ & $1.0$ & $3.7$ & $3.8$ & $3.9$ \\
		NMOS       & $94$ & $109$ & $109$ & $42$ & $51$ & $51$ & $9$ & $9$ & $9$ & $0.5$ & $0.6$ & $0.6$ & $5.5$ & $5.5$ & $5.6$ \\
		PMOS       & $94$ & $109$ & $109$ & $43$ & $52$ & $52$ & $10$ & $10$ & $10$ & $0.5$ & $0.6$ & $0.6$ & $6.5$ & $6.5$ & $6.6$ \\
		Resistor        & $357$ & $522$ & $682$ & $228$ & $228$ & $228$ & $8$ & $8$ & $8$ & $3.9$ & $5.8$ & $7.5$ & $5.9$ & $5.9$ & $6.0$ \\
		\midrule
		\multicolumn{16}{l}{Base Circuits} \\
		Ahuja OTA     & $2\,404$ & $3\,963$ & $6\,149$ & $6\,751$ & $7\,307$ & $7\,949$ & $13$ & $13$ & $14$ & $0.8$ & $1.4$ & $1.6$ & $4.3$ & $4.6$ & $5.5$ \\
		Gate Driver    & $2\,055$ & $2\,387$ & $2\,831$ & $4\,402$ & $4\,818$ & $5\,315$ & $13$ & $13$ & $13$ & $1.7$ & $2.2$ & $3.4$ & $4.5$ & $4.9$ & $5.3$ \\
		HPF        & $7\,693$ & $10\,481$ & $13\,752$ & $4\,415$ & $4\,755$ & $5\,225$ & $14$ & $14$ & $15$ & $1.2$ & $1.6$ & $2.1$ & $3.5$ & $3.7$ & $3.9$ \\
		LDO        & $2\,877$ & $4\,401$ & $6\,771$ & $2\,412$ & $2\,597$ & $2\,776$ & $14$ & $14$ & $14$ & $1.0$ & $1.2$ & $2.0$ & $3.8$ & $4.0$ & $4.4$ \\
		LPF        & $8\,171$ & $10\,852$ & $14\,326$ & $4\,302$ & $4\,624$ & $5\,004$ & $14$ & $14$ & $15$ & $0.8$ & $1.3$ & $2.0$ & $3.5$ & $3.6$ & $3.8$ \\
		Miller OTA    & $5\,882$ & $8\,467$ & $10\,640$ & $4\,105$ & $4\,469$ & $4\,869$ & $14$ & $14$ & $15$ & $1.0$ & $1.4$ & $1.8$ & $3.6$ & $3.8$ & $4.0$ \\
		\bottomrule
	\end{tabular}
	}
\end{table}
\subsection{Additional Evaluation Analyses}
\label{sec:additional_evaluation}
This subsection documents the released task records and evaluation implementation, together with detailed answer-option construction and Task-B class-balance statistics. Deterministic validation, LLM-judge agreement, and open-ended evaluation are reported in Section~\ref{ssec:evaluation_robustness}.

\subsubsection{Released Data and Evaluation Implementation}
\label{ssec:released_data}
\begin{figure}[t]
	\centering
	\includegraphics[width=0.3\linewidth]{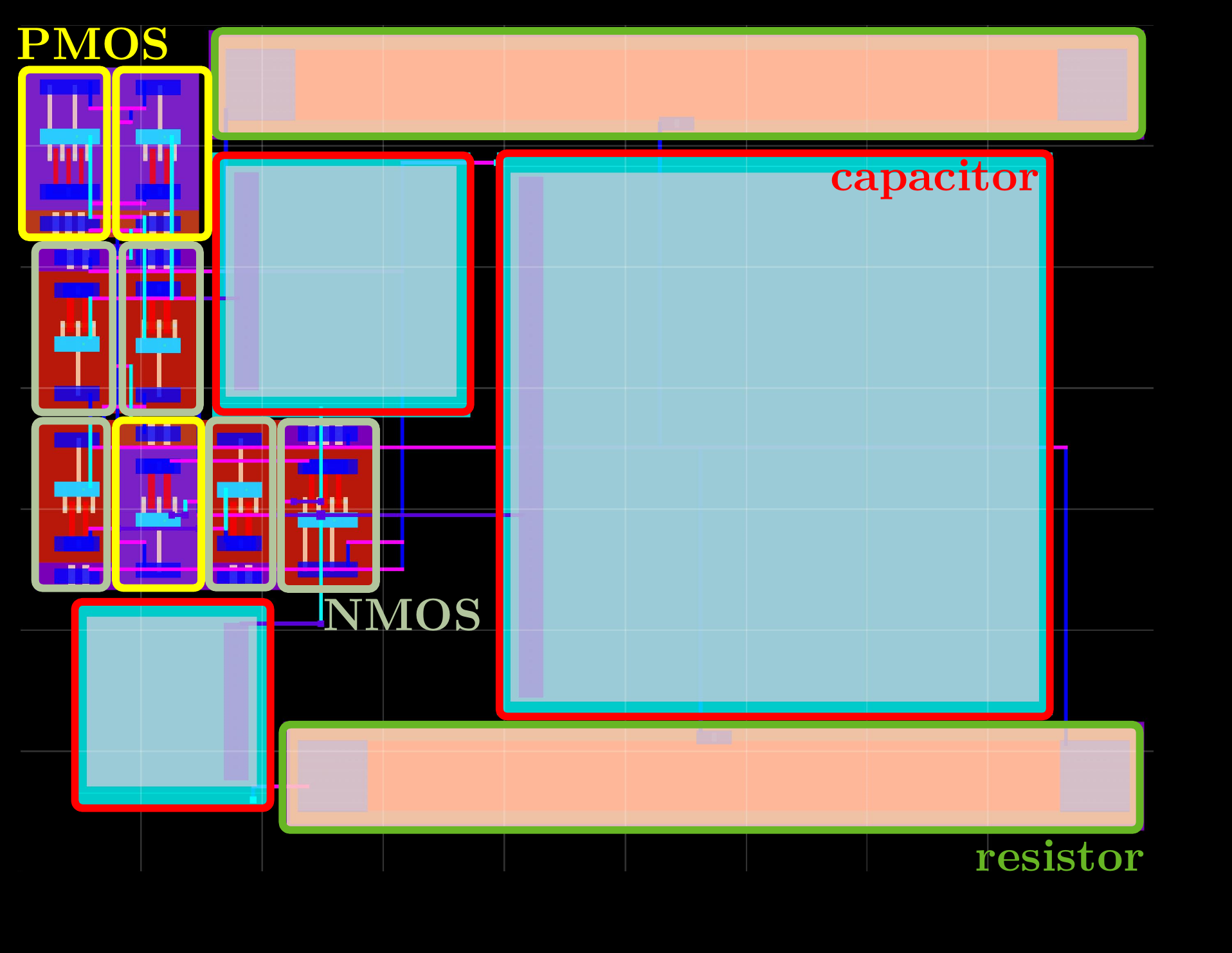}
	\caption{Example of a base circuit layout, an HPF. Devices are highlighted in coloured boxes.}
	\label{fig:layout_example}
\end{figure}

\begin{figure}[t]
	\centering
	\begin{subfigure}[b]{0.32\textwidth}
		\centering
		\includegraphics[width=\linewidth]{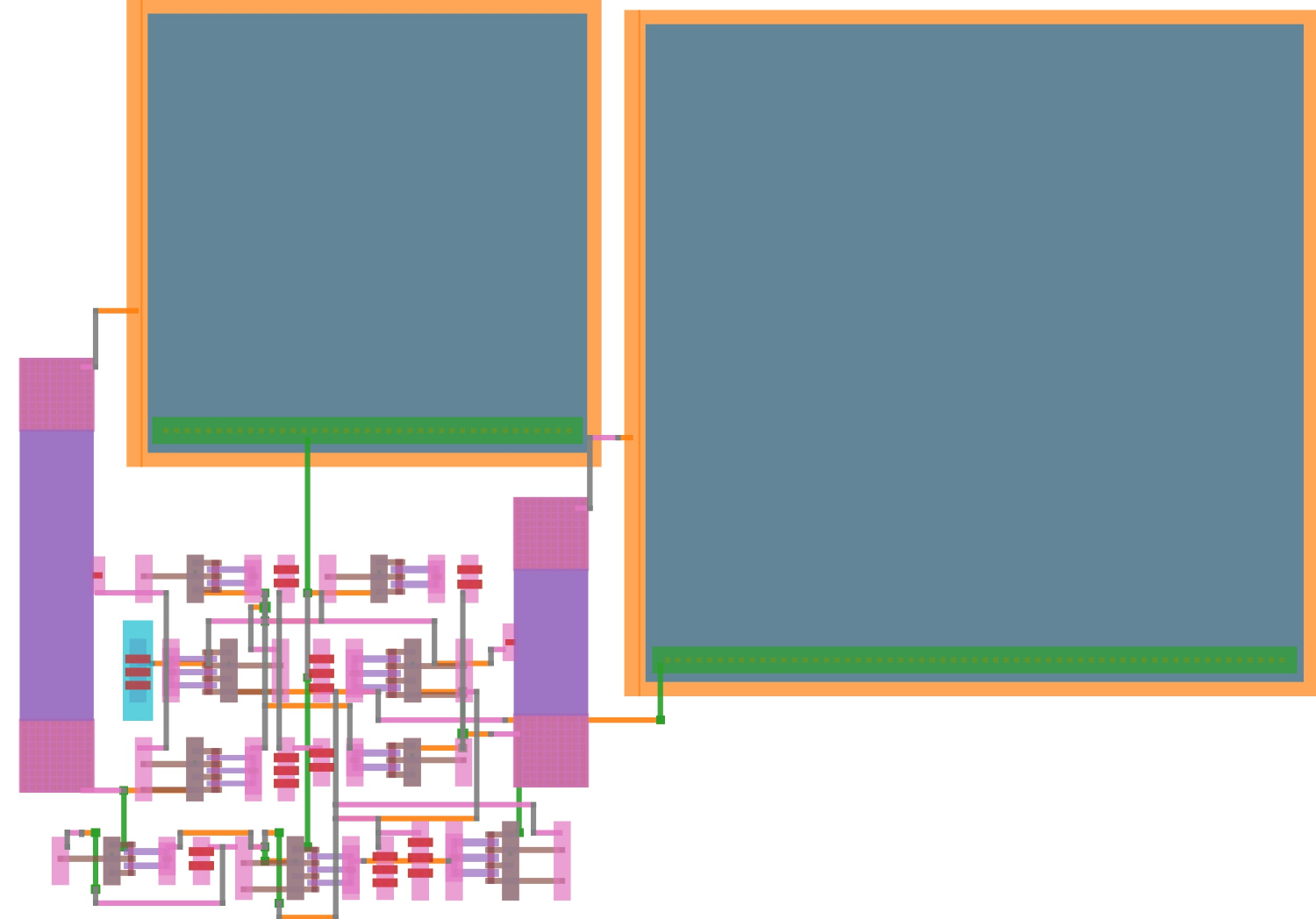}
		\caption{}
		\label{sfig:miller_ota_0}
	\end{subfigure}
	\hfill
	\begin{subfigure}[b]{0.32\textwidth}
		\centering
		\includegraphics[width=\linewidth]{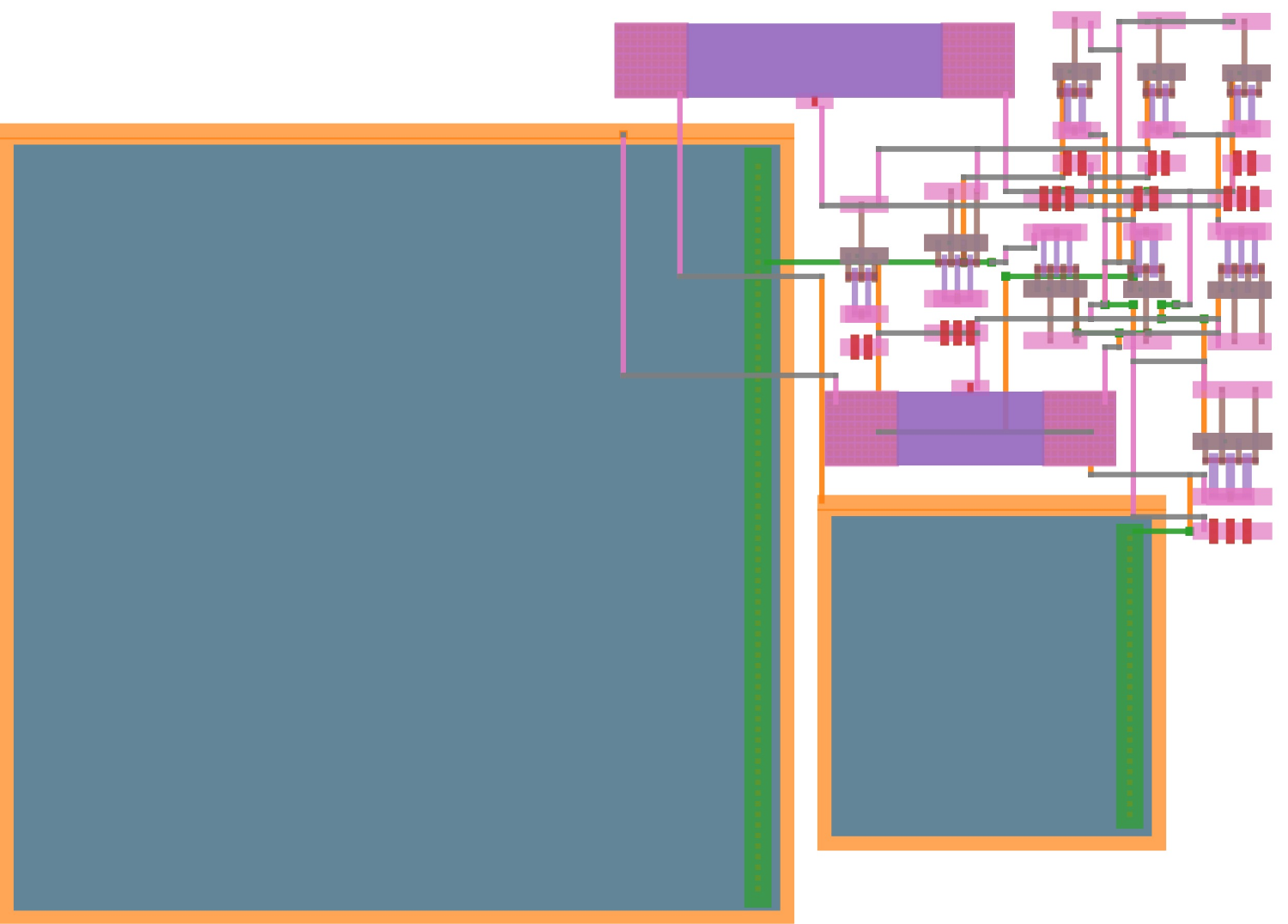}
		\caption{}
		\label{sfig:miller_ota_204}
	\end{subfigure}
	\hfill
	\begin{subfigure}[b]{0.32\textwidth}
		\centering
		\includegraphics[width=\linewidth]{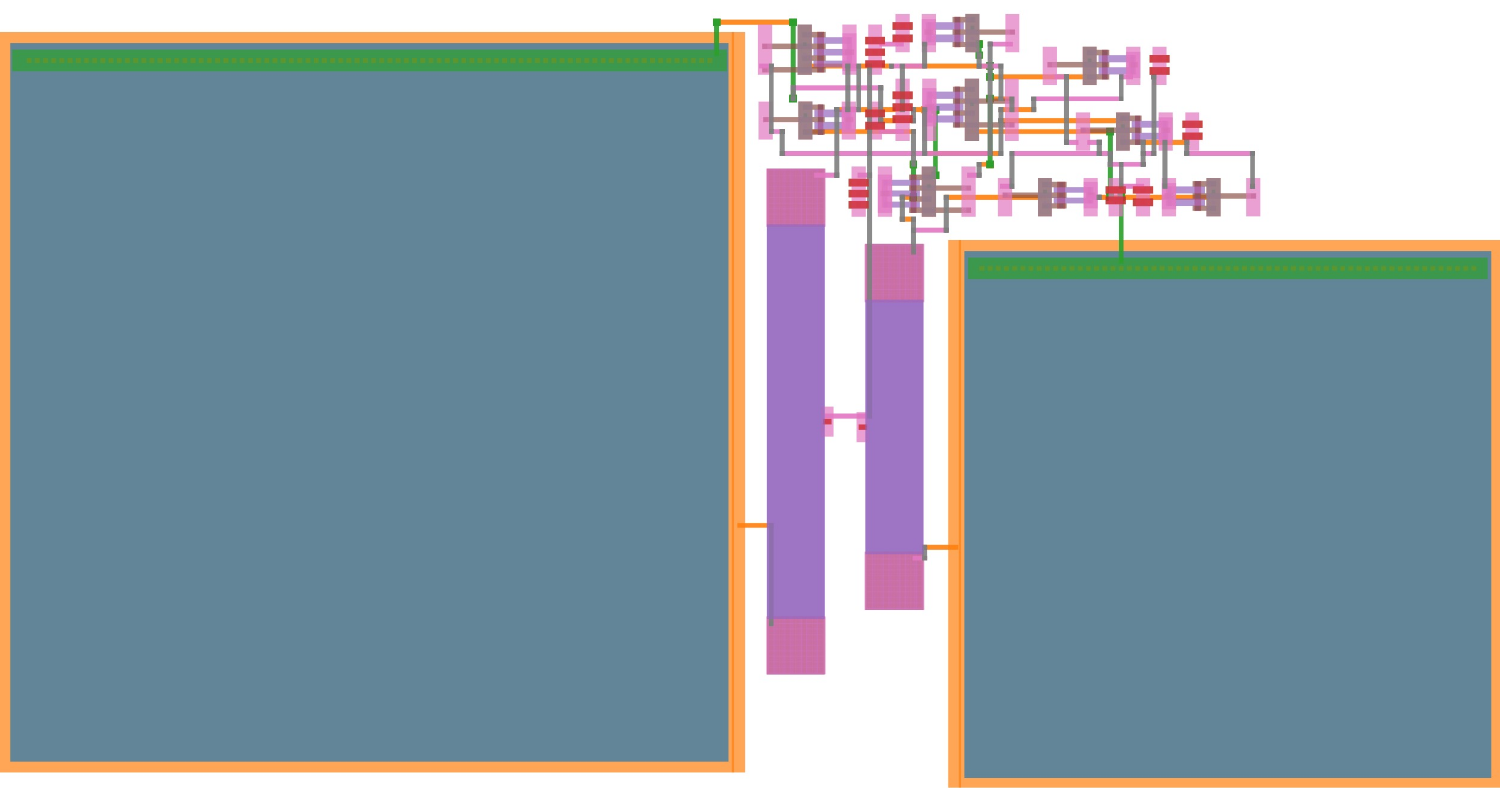}
		\caption{}
		\label{sfig:miller_ota_890}
	\end{subfigure}
	\caption{Examples of Miller OTA layouts.}
	\label{fig:miller_ota}
\end{figure}

\begin{figure}[t]
	\centering
	\begin{subfigure}[b]{0.32\textwidth}
		\centering
		\includegraphics[width=\linewidth]{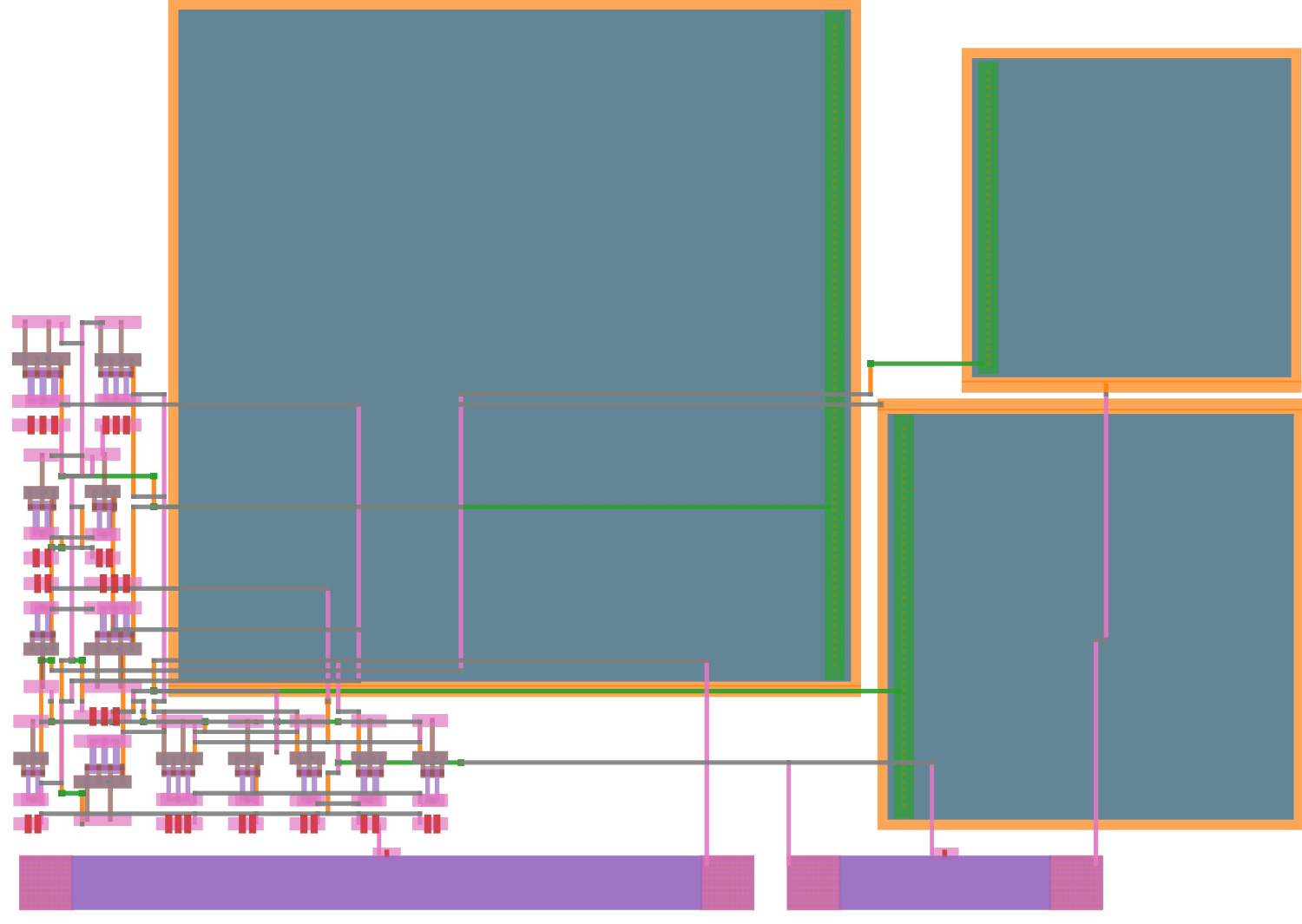}
		\caption{}
		\label{sfig:mixed_113}
	\end{subfigure}
	\hfill
	\begin{subfigure}[b]{0.32\textwidth}
		\centering
		\includegraphics[width=\linewidth]{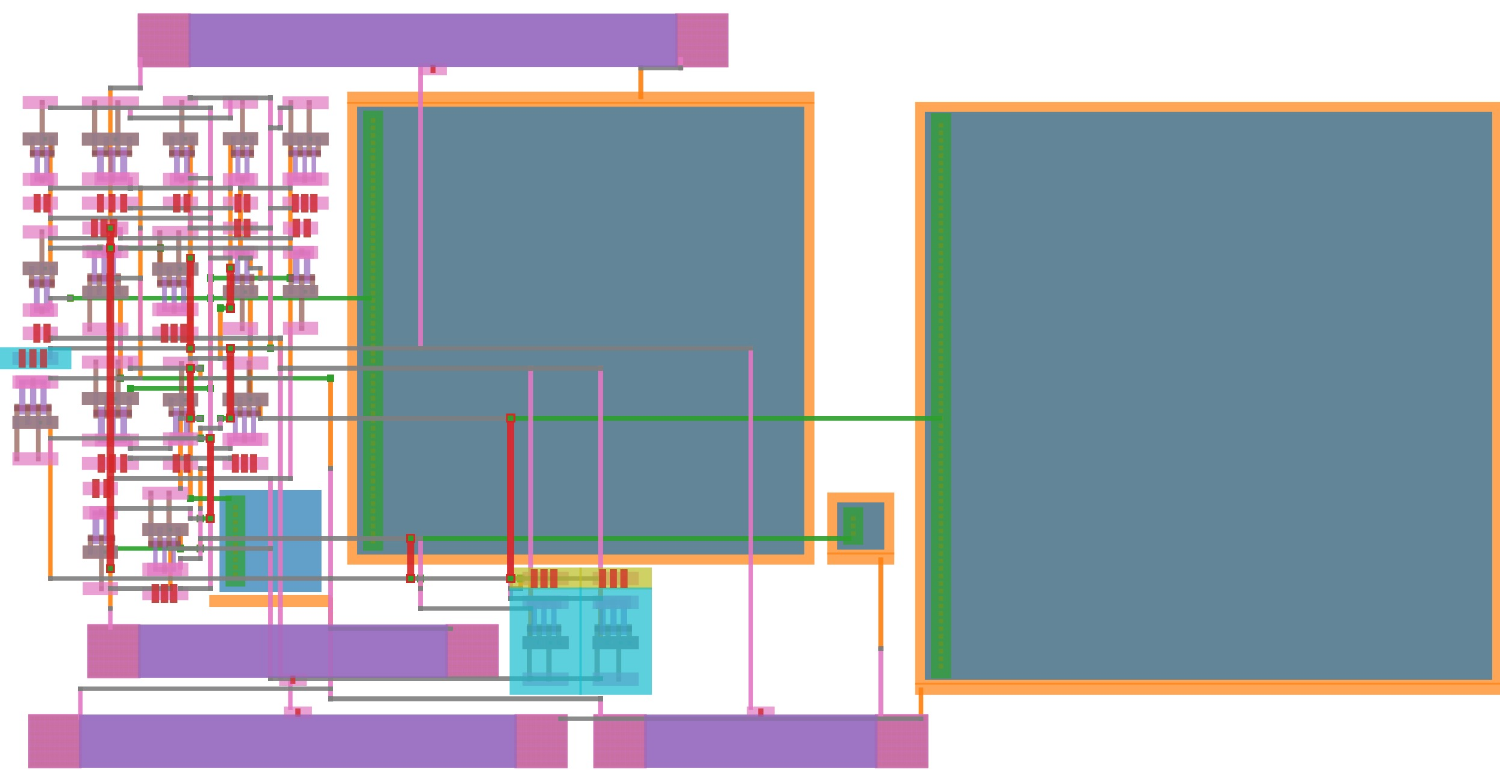}
		\caption{}
		\label{sfig:mixed_393}
	\end{subfigure}
	\hfill
	\begin{subfigure}[b]{0.32\textwidth}
		\centering
		\includegraphics[width=\linewidth]{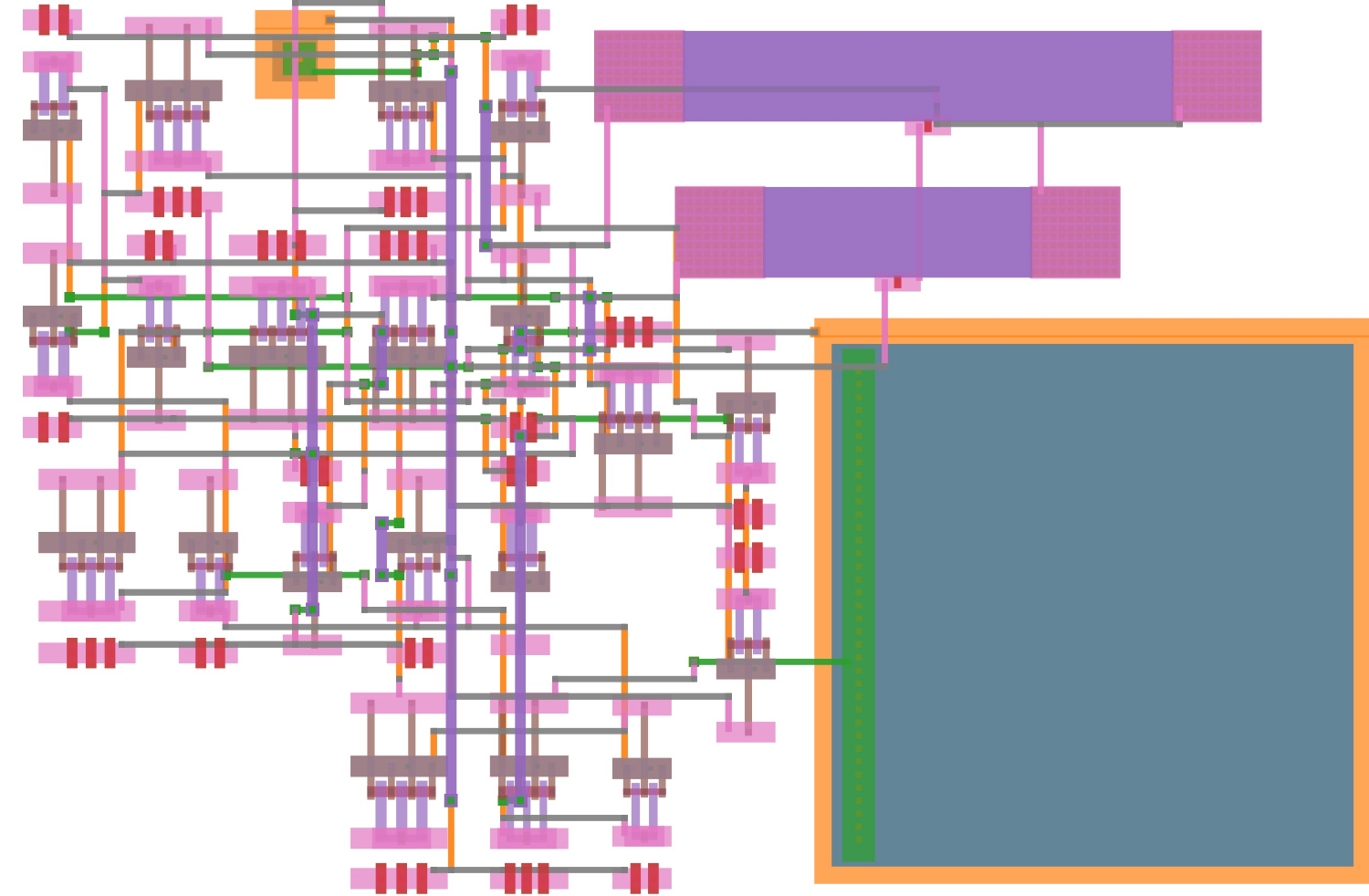}
		\caption{}
		\label{sfig:mixed_1454}
	\end{subfigure}
	\caption{Examples of mixed topologies.}
	\label{fig:mixed}
\end{figure}

The released repository contains the five task-specific train/validation/test JSON files, each pairing a relative PNG path with a two-turn user/assistant conversation. It also ships the InternLM LoRA fine-tuning launcher and preprocessing code, deterministic choice scoring, and the LLM-as-a-judge implementation. The deterministic scorer removes the image placeholder, extracts a final answer letter from the allowed A-F choices, and compares it with the ground-truth letter. The judge script stores the prompt, ground truth, raw prediction, parsed prediction, binary verdict, and disagreement statistics. It uses a model-agnostic plain-text prompt and deterministic decoding (temperature $0$, top-$p$ $1$, at most six generated tokens).
Table~\ref{tbl:sample_qas} gives one held-out record from each task. The examples show the released answer format and the variable number of answer options.
\begin{table*}[t]
	\centering
	\caption{Representative held-out Q\&As retrieved from the released THEIA task files. The answer letter is the assistant target in the corresponding JSON record.}
	\label{tbl:sample_qas}
	\scriptsize
	\renewcommand{\arraystretch}{1.12}
	\begin{tabular}{clp{0.58\textwidth}c}
	\toprule
	\textbf{Task} & \textbf{Layout} & \textbf{Question and choices} & \textbf{Target} \\
	\midrule
	A & \texttt{single\_cap\_131} & What type of device is this? A. NMOS; B. PMOS; C. resistor; D. capacitor. Pick one answer only. & D (capacitor) \\
	B & \texttt{ahuja\_ota\_481} & What circuit is shown? A. gate driver; B. LDO; C. Miller OTA; D. HPF; E. Ahuja OTA; F. LPF. Select exactly one option. & E (Ahuja OTA) \\
	C & \texttt{ahuja\_ota\_308} & Can you count the transistors in this circuit? A. 12; B. 14; C. 11; D. 15. Select exactly one option. & B (14) \\
	D & \texttt{mixed\_8977} & Can you count the PMOS devices? A. 10; B. 9; C. 8. Select exactly one option. & C (8) \\
	E & \texttt{mixed\_661} & What base circuits are combined? A--F enumerate topology multisets; the correct option is F: one HPF, one LDO, and one Miller OTA. Select exactly one option. & F \\
	\bottomrule
	\end{tabular}
\end{table*}
For the multiple-choice counting tasks C/D, the correct count is retained and distractors are sampled around it: nearby integers are used first, then valid random counts if needed. Candidate order is shuffled, and neither the number nor the position of answer options is fixed. Table~\ref{tbl:distractor_options} gives the resulting test-set option-count distributions.
\begin{table}[t]
	\centering
	\caption{Number of answer options for Task-C/D held-out questions after distractor construction.}
	\label{tbl:distractor_options}
	\scriptsize
	\begin{tabular}{lrrrr}
	\toprule
	\textbf{Task} & \textbf{3 options} & \textbf{4 options} & \textbf{5 options} & \textbf{6 options} \\
	\midrule
	C ($1\,399$) & $370$ & $359$ & $349$ & $321$ \\
	D ($1\,003$) & $272$ & $258$ & $223$ & $250$ \\
	\bottomrule
	\end{tabular}
\end{table}
\texttt{Task~B} is class-balanced: its six topology families occupy $16.3\%$-$17.1\%$ of every split~(Table~\ref{tbl:taskb_balance}). The majority-class predictor, always selecting Gate Driver, obtains $50/300=16.7\%$ on the test set, compared with $84\%$ for the fine-tuned VLM.
\begin{table}[t]
	\centering
	\caption{\texttt{Task-B} class counts (percentage within split) and majority-class baseline.}
	\label{tbl:taskb_balance}
	\scriptsize
	\resizebox{0.7\linewidth}{!}{%
	\begin{tabular}{lccc}
		\toprule
		\textbf{Topology} & \textbf{Train ($5\,302$)} & \textbf{Validation ($292$)} & \textbf{Test ($300$)} \\
		\midrule
		Ahuja OTA & $895$ ($16.9\%$) & $49$ ($16.8\%$) & $51$ ($17.0\%$) \\
		Gate Driver & $900$ ($17.0\%$) & $50$ ($17.1\%$) & $50$ ($16.7\%$) \\
		HPF & $865$ ($16.3\%$) & $48$ ($16.4\%$) & $49$ ($16.3\%$) \\
		LDO & $890$ ($16.8\%$) & $49$ ($16.8\%$) & $50$ ($16.7\%$) \\
		LPF & $873$ ($16.5\%$) & $48$ ($16.4\%$) & $50$ ($16.7\%$) \\
		Miller OTA & $879$ ($16.6\%$) & $48$ ($16.4\%$) & $50$ ($16.7\%$) \\
		\bottomrule
	\end{tabular}}
\end{table}
Figure~\ref{fig:layout_example} shows a high-pass filter~(HPF) layout, where each device is highlighted by a colourful box.
Variability among circuit variants present in the dataset is exemplified in Figure~\ref{fig:miller_ota} and Figure~\ref{fig:mixed}, which show multiple Miller OTA and Low-Dropout Regulator layouts, respectively.

\clearpage


\newpage
\section*{NeurIPS Paper Checklist}

\begin{enumerate}

\item {\bf Claims}
    \item[] Question: Do the main claims made in the abstract and introduction accurately reflect the paper's contributions and scope?
    \item[] Answer: \answerYes{} 
    \item[] Justification: The main contributions of the paper are clearly stated both in the Abstract and in the Introduction~(Section~\ref{sec:introduction}), where a dedicated portion of it discusses the main contributions of this work and how they are positioned with respect to the state of the art.
    \item[] Guidelines:

\item {\bf Limitations}
    \item[] Question: Does the paper discuss the limitations of the work performed by the authors?
    \item[] Answer: \answerYes{} 
    \item[] Justification: The limitations are discussed in a dedicated section, see Section~\ref{sec:limitations}.

\item {\bf Theory assumptions and proofs}
    \item[] Question: For each theoretical result, does the paper provide the full set of assumptions and a complete (and correct) proof?
    \item[] Answer: \answerNA{} 
    \item[] Justification: This work does not provide any novel theoretical result such as new theorems or formulas.

    \item {\bf Experimental result reproducibility}
    \item[] Question: Does the paper fully disclose all the information needed to reproduce the main experimental results of the paper to the extent that it affects the main claims and/or conclusions of the paper (regardless of whether the code and data are provided or not)?
    \item[] Answer: \answerYes{} 
    \item[] Justification: This work introduces a novel dataset aimed at advancing analog layouts analysis ML-driven techniques. It includes all the necessary information to assess the data creation process as well as to reproduce the VLM fine-tuning baseline results. Details are reported in Section~\ref{sec:methodology}, Section~\ref{sec:results}, as well as in Appendix~\ref{sec:a_appendix}.

\item {\bf Open access to data and code}
    \item[] Question: Does the paper provide open access to the data and code, with sufficient instructions to faithfully reproduce the main experimental results, as described in supplemental material?
    \item[] Answer: \answerYes{} 
    \item[] Justification: The main contribution of this work is a novel dataset aimed at advancing analog layouts analysis ML-driven techniques. The dataset is released open-source as well as the code used in the baseline methodology for VLM fine-tuning and evaluation.

\item {\bf Experimental setting/details}
    \item[] Question: Does the paper specify all the training and test details (e.g., data splits, hyperparameters, how they were chosen, type of optimizer) necessary to understand the results?
    \item[] Answer: \answerYes{} 
    \item[] Justification: This work includes all the necessary information to assess the data creation process as well as to reproduce the VLM fine-tuning baseline results. Details are reported in Section~\ref{sec:methodology}, Section~\ref{sec:results}, as well as in Appendix~\ref{sec:a_appendix}.

\item {\bf Experiment statistical significance}
    \item[] Question: Does the paper report error bars suitably and correctly defined or other appropriate information about the statistical significance of the experiments?
    \item[] Answer: \answerNo{} 
    \item[] Justification: The main contribution of this work is a novel dataset aimed at advancing analog design optimization techniques targeting large-scale topologies, not the baseline VLM fine-tuning technique. The dataset is released open-source and it is fully characterized by statistical measures.

\item {\bf Experiments compute resources}
    \item[] Question: For each experiment, does the paper provide sufficient information on the computer resources (type of compute workers, memory, time of execution) needed to reproduce the experiments?
    \item[] Answer: \answerYes{} 
    \item[] Justification: Section~\ref{sec:results} report the computer resources employed to generate the dataset and carry out the experiments, respectively.
    
\item {\bf Code of ethics}
    \item[] Question: Does the research conducted in the paper conform, in every respect, with the NeurIPS Code of Ethics \url{https://neurips.cc/public/EthicsGuidelines}?
    \item[] Answer: \answerYes{} 
    \item[] Justification: The research conducted in this paper conforms, in every respect, with the NeurIPS Code of Ethics.

\item {\bf Broader impacts}
    \item[] Question: Does the paper discuss both potential positive societal impacts and negative societal impacts of the work performed?
    \item[] Answer: \answerYes{} 
    \item[] Justification: The main contributions of the paper are clearly stated both in the Abstract and in the Introduction~(Section~\ref{sec:introduction}). Moreover, Related Work~(Section~\ref{sec:background}) and Conclusions~(Section~\ref{sec:conclusions}) discuss how THEIA can foster research toward new frontiers of analog circuit design automation.
    
\item {\bf Safeguards}
    \item[] Question: Does the paper describe safeguards that have been put in place for responsible release of data or models that have a high risk for misuse (e.g., pre-trained language models, image generators, or scraped datasets)?
    \item[] Answer: \answerNA{} 
    \item[] Justification: The proposed dataset (and baseline) does not pose significant threats of misuse.

\item {\bf Licenses for existing assets}
    \item[] Question: Are the creators or original owners of assets (e.g., code, data, models), used in the paper, properly credited and are the license and terms of use explicitly mentioned and properly respected?
    \item[] Answer: \answerYes{} 
    \item[] Justification: All the people involved in the creation, curation, and diffusion of the assets mentioned, used, or proposed in this work are properly mentioned and credited.

\item {\bf New assets}
    \item[] Question: Are new assets introduced in the paper well documented and is the documentation provided alongside the assets?
    \item[] Answer: \answerYes{} 
    \item[] Justification: All the people involved in the creation, curation, and diffusion of the assets mentioned, used, or proposed in this work are properly mentioned and credited.

\item {\bf Crowdsourcing and research with human subjects}
    \item[] Question: For crowdsourcing experiments and research with human subjects, does the paper include the full text of instructions given to participants and screenshots, if applicable, as well as details about compensation (if any)? 
    \item[] Answer: \answerNA{} 
    \item[] Justification: This work does not involve crowdsourcing nor research with human subjects.

\item {\bf Institutional review board (IRB) approvals or equivalent for research with human subjects}
    \item[] Question: Does the paper describe potential risks incurred by study participants, whether such risks were disclosed to the subjects, and whether Institutional Review Board (IRB) approvals (or an equivalent approval/review based on the requirements of your country or institution) were obtained?
    \item[] Answer: \answerNA{} 
    \item[] Justification:  This work does not involve crowdsourcing nor research with human subjects.

\item {\bf Declaration of LLM usage}
    \item[] Question: Does the paper describe the usage of LLMs if it is an important, original, or non-standard component of the core methods in this research? Note that if the LLM is used only for writing, editing, or formatting purposes and does \emph{not} impact the core methodology, scientific rigor, or originality of the research, declaration is not required.
    \item[] Answer: \answerYes{} 
    \item[] Justification: LLMs are employed in the methodology. Section~\ref{sec:methodology} and Section~\ref{sec:results} discusses the role of LLMs in the methodology and experimental validation, respectively. LLMs were not involved in the concept development of the dataset nor the baseline methodology.

\end{enumerate}

\end{document}